\pdfoutput=1
\documentclass[11pt]{article}

\usepackage[final]{acl}

\usepackage{times}
\usepackage{latexsym}
\usepackage{enumitem}
\usepackage{longtable}
\usepackage{multicol}
\usepackage{adjustbox}
\usepackage{mathtools}
\usepackage[T1]{fontenc}
\usepackage[utf8]{inputenc}

\usepackage{microtype}

\usepackage{inconsolata}

\usepackage{graphicx}
\usepackage{amsmath}
\usepackage{xcolor}
\definecolor{darkred}{RGB}{139,0,0}

\usepackage{multirow}
\usepackage{tabularx}
\usepackage{booktabs}
\usepackage{array}
\usepackage{paracol}
\title{Rank-Then-Score: Enhancing Large Language Models for \\ Automated Essay Scoring}

\author{
  \textbf{Yida Cai\textsuperscript{1,2}}, 
  \textbf{Kun Liang\textsuperscript{1,4}}, 
  \textbf{Sanwoo Lee\textsuperscript{1,3}}, 
  \textbf{Qinghan Wang\textsuperscript{3}}, 
  \textbf{Yunfang Wu\textsuperscript{1,3}}
  \\
  \textsuperscript{1}MOE Key Laboratory of Computational Linguistics, Peking University \\
  \textsuperscript{2}School of Software and Microelectronics, Peking University \\
 \textsuperscript{3}School of Computer Science, Peking University \\
 \textsuperscript{4}School of Artificial Intelligence, Beijing Normal University\\
  \texttt{caiyida@stu.pku.edu.cn, 202111081003@mail.bnu.edu.cn,\{wuyf,sanwoo\}@pku.edu.cn}
}

\begin{document}
\maketitle
\begin{abstract}
%The progress of Automatic Essay Scoring (AES) always rely on advancements in models. 
In recent years, large language models (LLMs) achieve remarkable success across a variety of tasks. However, their potential in the domain of Automated Essay Scoring (AES) remains largely underexplored. Moreover, compared to English data, the methods for Chinese AES is not well developed. In this paper, we propose Rank-Then-Score (RTS), a fine-tuning framework based on large language models to enhance their essay scoring capabilities. Specifically, we fine-tune the ranking model (Ranker) with feature-enriched data, and then feed the output of the ranking model, in the form of a candidate score set, with the essay content into the scoring model (Scorer) to produce the final score. Experimental results on two benchmark datasets, HSK and ASAP, demonstrate that RTS consistently outperforms the direct prompting (Vanilla) method in terms of average QWK across all LLMs and datasets, and achieves the best performance on Chinese essay scoring using the HSK dataset.
\end{abstract}

\section{Introduction}
Automated Essay Scoring (AES) is a task that uses machine learning methods to score an essay, which shows great efficiency and objectivity compared to humans \cite{dikli2006overview}. The traditional \textit{prompt-specific} AES task (Prompt means Topic) focuses on essays within the same prompt, allowing the scoring model to more accurately capture the scoring criteria for that specific prompt. As a result, it aligns more closely with human scoring criteria and better meets the precision requirements for large-scale examinations \cite{attali2006automated}.

Previous research primarily focus on modeling the content of essays using neural network models \cite{taghipour2016neural,dong2017attentionbased}. Subsequently, researchers 
%aimed 
try to explore the enhancement of performance by modeling various types of content-related information. Some works achieve satisfactory results in both cross-prompt and prompt-specific tasks by modeling statistical features \cite{ridley2020prompt}. Some studies achieve significant improvements, even reaching state-of-the-art (SOTA) levels, by incorporating ranking tasks into the scoring process \cite{yang2020enhancing,xie2022automated}. Compared to the absolute quality represented by scoring, ranking can reflect differences between essays through relative quality, thereby reducing 
%errors 
biases caused by absolute scoring. This idea is also commonly used in the optimization process of Reward Models in Reinforcement Learning from Human Feedback (RLHF) \cite{li2024processrewardmodelqvalue}. 

In recent years, with the advancement of Large Language Models (LLMs), many text regression tasks and text evaluation tasks witness further progress \cite{vacareanu2024from,chen2023exploring}.
% In the field of AES, some studies have initially investigated the zero-shot capabilities of LLMs, exploring several viable approaches \cite{lee2024unleashing}.
Some studies explore AES methods based on fine-tuning, which achieve considerable improvements compared to zero-shot approaches \cite{stahl2024exploring,li2024conundrums}. 

However, existing methods that combine ranking with regression face a key challenge in the LLM era: the training logic of LLMs(focused on next-token prediction) is not directly compatible with regression or ranking loss formulations \cite{yang2020enhancing}. Consequently, these methods struggle to transfer and leverage the full advantages of LLMs, such as their powerful semantic understanding and multi-task generalization capabilities. Meanwhile, some studies consider the integration of ranking task with the scoring task, but the results of such integration do not reach a level comparable to that of supervised small models, and further exploration is needed \cite{stahl2024exploring}.

% From an overall perspective, most AES works based on LLMs focus on improving the performance of a single model on a single task through specific instructions, without fully leveraging the characteristic of LLMs that they generally perform well across different tasks. 

%这里的ceiling of llms是否合适
%from cai：我觉得是有点夸张，改成了与小模型supervised相当的水平

Additionally, most current AES research focuses on the English language. However, when writing essays in different languages, such as Chinese, the evaluation criteria can vary significantly. Therefore, some researchers also explore AES methods in Chinese essays and achieve certain progress \cite{song2020hierarchical,song2020multistage,he2022automated}.

In this paper, we propose a novel Rank-Then-Score (RTS) pipeline, which leverages two key advantages: (1) Our multi-stage design (ranking → scoring) decouples complex tasks into manageable sub-problems, aligning with LLMs' strength in specialized fine-tuning; (2) Building on the success of LLMs as rankers in recommendation systems \cite{hou2024large} and RLHF \cite{bai2022training}, our Ranker effectively narrows score intervals through pairwise comparisons, ensuring more accurate predictions.

Specificly, RTS comprises two models in two different tasks: \textbf{a Ranker and a Scorer}. We first randomly selected reference essays and incorporated essay-related features into the essays. Then, we employ a fine-tuned pairwise Ranker to compare the target essay with the reference essays in a manner analogous to a binary search tree. This process identifies a candidate set of scores for the target essay. Finally, the candidate score set, along with the target essay, is fed into the fine-tuned Scorer to obtain the predicted score. 

Moreover, this paper also employs a new Chinese essay scoring dataset: the HSK dataset, which is derived from the essay data of the HSK examination \cite{yong2022}. We clean and filter the data, ultimately obtaining a dataset comprising 8,597 essays.

We conduct experiments on Chinese (HSK) and English (ASAP) \cite{asap-aes} datasets using various LLMs for ranking and scoring tasks. The fine-tuned RTS method is compared to a Vanilla baseline fine-tuned with standard instructions, and Quadratic Weighted Kappa (QWK) is used as the evaluation metric. As shown in the Table \ref{tab:hsk} and Table \ref{tab:asap}, the RTS method outperforms the Vanilla method across all settings. Specifically, on the HSK dataset, RTS achieves an improvement of 1.9\% ${(74.6\%\to76.5\%)}$ 
%1.9\% {(74.6\% \to 76.5\%)}
over the Vanilla method, while on the ASAP dataset, it achieves improvements of 1.7\% ${(78.1\% \to 79.8\%)}$ and 1.1\% ${(78.3\% \to 79.4\%)}$ over the Vanilla method in different configurations. Additionally, on the HSK dataset, RTS surpasses other methods using smaller models. On the ASAP dataset, RTS is comparable to the R${^2}$BERT \cite{yang2020enhancing} method and approaches the performance of the NPCR \cite{xie2022automated} method.

In summary, our contributions are as follows:
\begin{itemize}
  \item We propose a method for integrating ranking and scoring 
  %tasks 
  mechanisms within LLMs, thereby enhancing the performance of LLMs for AES.
  %in the AES task.
  \item We propose a \textbf{Binary-Search-Tree}-like approach to transform the results of pairwise ranking into inputs for the scoring model.
  \item We present a novel method that incorporates rich essay-related \textbf{features} into the scoring task.
  %\item We filter and utilize a new Chinese AES dataset: \textbf{HSK}.
\end{itemize}

\section{Related Work}
\textbf{Automated Essay Scoring} The development of AES is mainly driven by technological advancements and researchers' exploration of essay evaluation criteria. Early methods primarily relied on hand-crafted features \cite{yannakoudakis2011dataset,persing2013modeling}. Subsequently, many studies began to introduce neural network models and achieved excellent results \cite{taghipour2016neural,dong2017attentionbased,farag2018neural}. At the same time, methods that utilized features \cite{ridley2020prompt,chen2023pmaes} and ranking \cite{yang2020enhancing,xie2022automated} also emerged. In recent years, an increasing number of studies focus on Multi-Trait Scoring methods \cite{ridley2021automated,li2024conundrums}, which are widely applied in various essay scoring works.

After the emergence of LLMs, many researchers believed that the characteristic of LLMs performing well across various downstream tasks is worth leveraging for the AES task. Among them, the work of  \cite{lee2024unleashing} explored the performance of the Multi-Trait method in a zero-shot setting on LLMs, while \cite{xiao2024humanai} investigated the potential of fine-tuning LLMs to scoring. Recently, \cite{stahl2024exploring} explores various instruction methods in the in-context learning of LLMs, achieving comprehensive results in this field. 

\begin{figure*}[ht]
    \centering
    \includegraphics[width=0.8\textwidth,height=0.33\textheight]{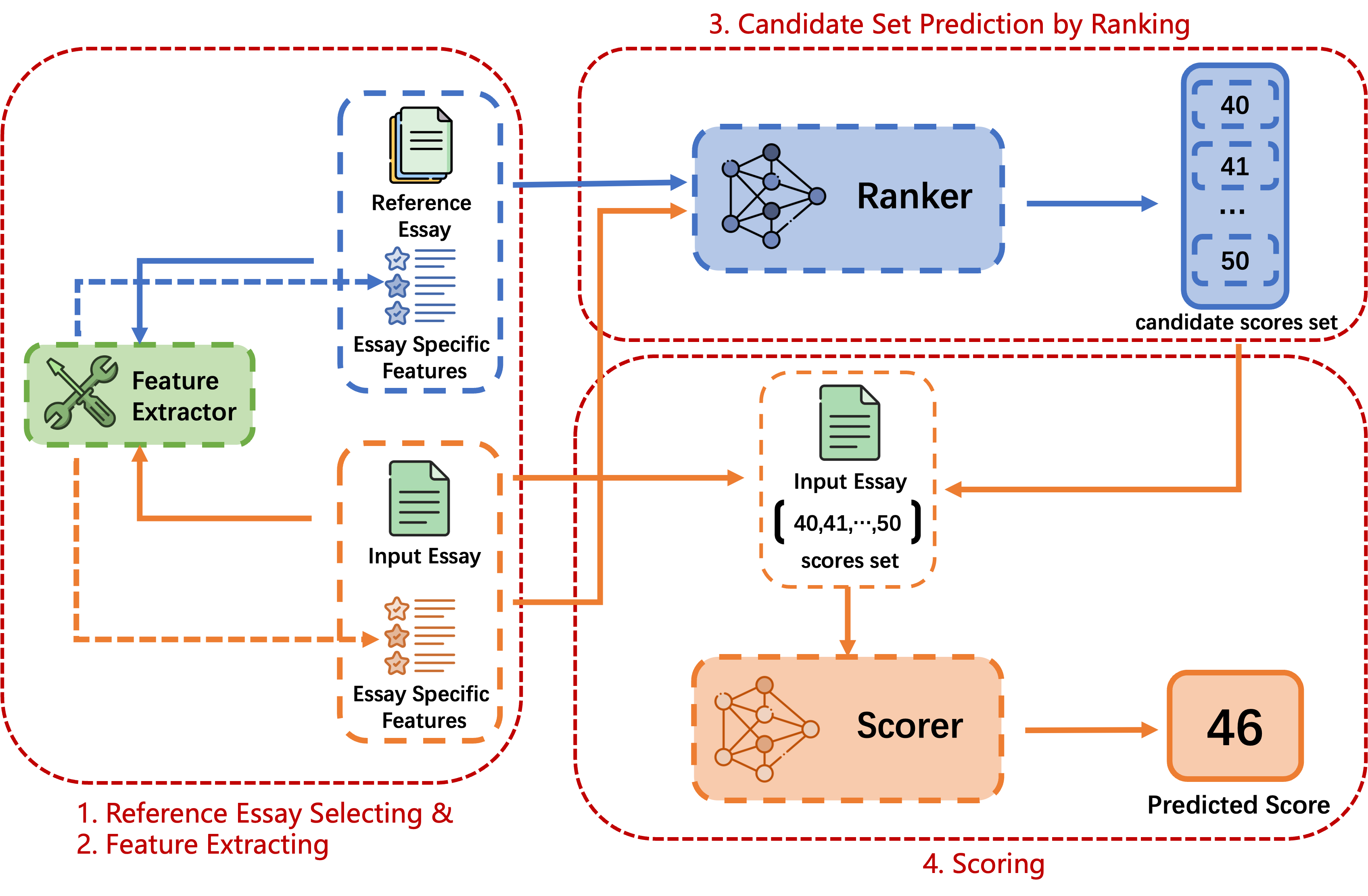}
    \caption{The overall architecture of RTS is illustrated in the figure. Excluding the training process, the method is divided into the following four steps: (1) Select reference essays. (2) use the feature extractor to identify the features of the essays, and incorporate these features into the essay content. (3) Utilize the Ranker to obtain the candidate score set of the current essay through pairwise ranking. (4) Feed the candidate score set, along with the essay, into the Scorer to generate final score.}
    \label{fig:architecture}
\end{figure*}

\textbf{Chinese AES} In addition to these developments, there are also some advanced explorations in the field of Chinese AES. Firstly, in the context of pre-trained methods, research on Chinese AES also directs its approach towards Multi-Trait Scoring \cite{song2020hierarchical,song2020multistage}. Moreover, \cite{gong2021iflyea} meticulously listed the majority of aspects that need to be considered in Chinese AES, providing significant guidance for future research. Following that, \cite{he2022automated} proposed a new method based on multiple scorers, which achieved considerable improvement.

However, it is unfortunate that there is a scarcity of research on Chinese AES based on LLMs, which is also the direction we are striving towards.

\section{Method}
The supervised fine-tuning-based AES method can be formalized as follows: given a set of essays $\mathcal{X} = \{x_1, x_2, \dots, x_n\}$ and a corresponding set of scores $\mathcal{Y} = \{y_1, y_2, \dots, y_n\}$, where each essay $x_i$ is associated with a ground truth score $y_i$. Given a pre-trained model $g_\theta$ that is typically parameterized by $\theta$. The goal is to train the base $g_\theta$ and obtain a new model $g_{\hat{\theta}}$ with $\hat{\theta}$ that predicts a score $\hat{y}_i = g_{\hat{\theta}}(x_i)$, making $\hat{y}_i$ close to $y_i$. 

The RTS method divides the scoring process into two steps: (1) Training a pairwise ranking model (Ranker) to generate candidate score sets for target essays. (2) Training a scoring model (Scorer) to predict the real scores. The architecture of RTS is illustrated in Figure \ref{fig:architecture}. 

\subsection{Pairwise Ranking}
The task for the Ranker is as follows: given a target essay and a reference essay, the model outputs the index of the essay that has the higher score; We repeat the process above and transform the ranking results into a candidate score set.

% In zero-shot scenarios, LLMs struggle to follow the output format specified in the prompt, and extracting reliable indicators of the better essay becomes challenging. Therefore, 

We employ supervised fine-tuning method on an LLM, allowing it to accurately evaluate the quality of essays through ranking. We design a four-step approach to train the Ranker's pairwise ranking capability and generate the candidate score set:
\begin{enumerate}
    \item \textbf{Reference Essay Selecting}: For each prompt, a subset of reference essays is selected to facilitate pairwise comparisons. 
    \item \textbf{Features Extracting}: This includes linguistic features, structural features, and semantic features to effectively represent the essays. 
    \item \textbf{Fine-tuning Pairwise Ranker}: We fine-tune the model using feature-augmented pairwise data. 
    \item \textbf{Candidate Set Prediction by Ranking}: By comparing the target essay with the reference essays, the model predicts the candidate score set for the target essay. 
\end{enumerate}

In Step 1, we select different reference scores for different prompts. 
% The selection of reference scores is primarily based on the upper and lower bounds of the essay scores and the number of score intervals in the candidate score set. 
Specifically, we adhere to the following two principles for selection:(1) the number of reference scores should not exceed 5, as exceeding this limit would increase inference costs. (2) when the number of scores is even, the two middle scores are selected; when the number is odd, the central score is selected. 
% The selected reference scores are shown in Table~\ref{tab:prompt_scores}. 
Afterwards, for each reference score, we randomly select \textbf{2 essays} as reference essays. 

In Step 2, we utilize features related to the text to enhance the model's understanding of the essay. We first extract various types of feature for both Chinese and English data. For the ASAP dataset, we use the hand-crafted features proposed by \cite{ridley2020prompt}. For the HSK dataset, we adopt the feature categories used by \cite{li2022unified} in their readability assessment study and extract features by ourselves. The specific feature categories are detailed in the Appendix~\ref{sec:appendixA}. 

Afterwards, we employ \textbf{LibSVM} to select a subset of beneficial features. Specifically, we use it to perform simple predictions on pairs \((f, y)\), where $f$ represents features and $y$ represents scores, the F-score is defined by \cite{chen2006combining} as:
\begin{equation}
F_i \equiv 
\frac{
    \left( \bar{f}_i^{(+)} - \bar{f}_i \right)^2 + \left( \bar{f}_i^{(-)} - \bar{f}_i \right)^2
}{
    \splitfrac{
        \tfrac{1}{n_+ - 1} \sum_{j=1}^{n_+} \left( f_{j,i}^{(+)} - \bar{f}_i^{(+)} \right)^2
    }{
        + \tfrac{1}{n_- - 1} \sum_{j=1}^{n_-} \left( f_{j,i}^{(-)} - \bar{f}_i^{(-)} \right)^2
    }
}
\end{equation}
where $\bar{f}_i, \bar{f}_i^{(+)}, \bar{f}_i^{(-)}$ are the average of the $i$th feature of the whole, positive, and negative data; $f_{j,i}^{(+)}$ is the $i$th feature of the $j$th positive instance, and $f_{j,i}^{(-)}$ is the $i$th feature of the $j$th negative instance. Then, we select the top 10 features with the highest F-score as the final set of features.
%\begin{equation}
%\text{Top10Features} = \underset{i \in \{1, 2, \dots, n\}}{\text{argtopk}} \, F_i, \quad k=10,
%\end{equation}
%where \( \text{argtopk} \) is the operator that returns the top \( k \) features with the highest F-scores. 
We concatenate the content of each essay with its features to obtain a feature-augmented essay representation, which serves as the input 
%the essay to be used 
in the future steps. 

In Step 3, in a training set of size \( M \), for each feature-enhanced essay, we randomly select \( k \) essays with different scores to form pairwise data, which is used to fine-tune the Ranker. The instruction used for fine-tuning on ASAP is shown as Figure \ref{fig:rank_instruction}. The instruction used on HSK is shown in Appendix \ref{sec:HSK}
\begin{figure}[t]
    \centering
    \includegraphics[width=0.8\columnwidth,height=0.26\textheight]{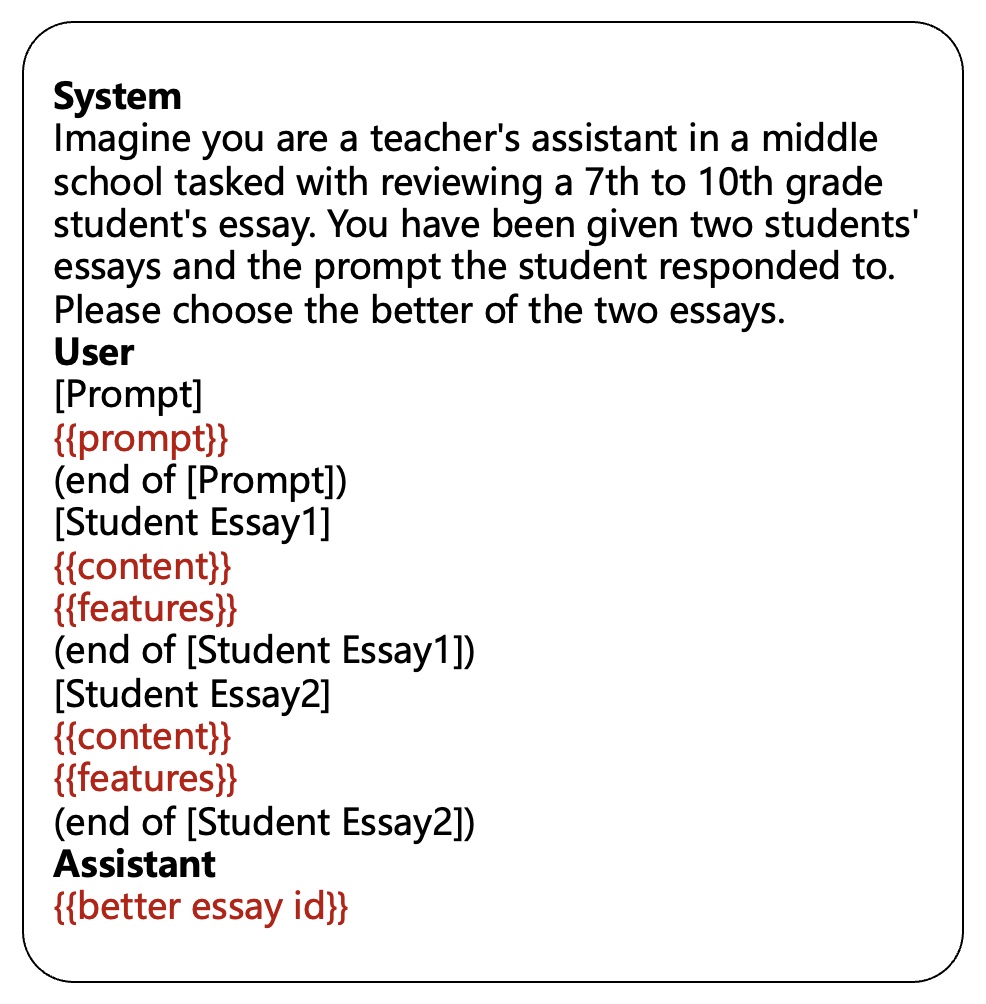}
    \caption{Instruction for fine-tuning the Ranker. Contents to be filled are highlighted in \textcolor{darkred}{red}.}
    \label{fig:rank_instruction}
\end{figure}

\begin{figure}[t]
    \centering
    \includegraphics[width=\columnwidth]{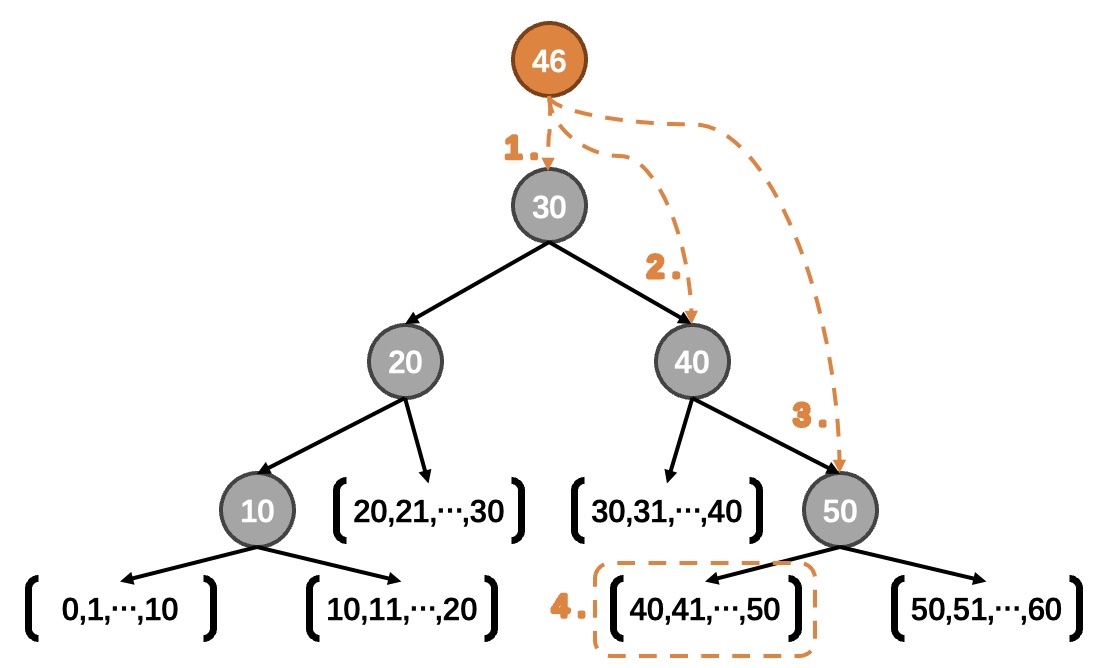}
    \caption{The BST-like inference process.}
    \label{fig:BST}
\end{figure}

\begin{figure}[h]
    \centering
    \includegraphics[width=0.5\columnwidth]{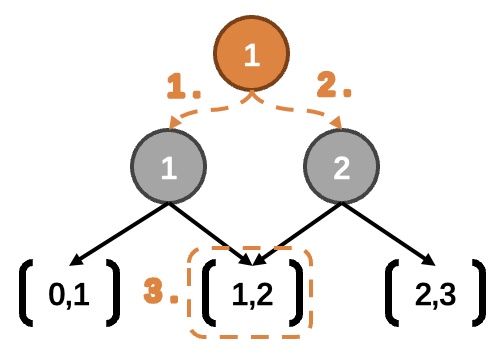}
    \caption{Another scenario of the BST-like approach.}
    \label{fig:BST2}
\end{figure}

In Step 4, we adopt a \textbf{"Binary Search Tree"} approach, inspired by \cite{zhuang2024setwise}. By using the Ranker, we compare the target essay with the reference essays to determine the candidate score set. The detailed process is illustrated in Figure \ref{fig:BST}. We arrange the reference essays in a BST structure and begin pairwise ranking from the score closest to the median. After each round, we use the result to guide the selection of the next score, ultimately obtaining a candidate score set represented by the leaf node. In special cases, when the Ranker determines that the target essay's score lies between two adjacent essays, we add an additional leaf node between the two reference essays to fix this issue as shown in Figure \ref{fig:BST2}. 

%We arrange the reference essays in a binary search tree structure and begin pairwise ranking from the score closest to the median. After each round, we use the result to guide the selection of the next score, ultimately obtaining a candidate score set represented by the leaf node.

%We also define a separate candidate score set between adjacent reference scores.

During each comparison with reference essays, we employ the following \textbf{Multi-Validation} method based on \cite{qin2023large} to assess the difference. Given two essays \( e_1 \) and \( e_2 \), we define the comparison function \( C(e_1, e_2) \) as follows:
\begin{equation}
C(e_1, e_2) = 
\begin{cases}
1, & \text{if } e_1 \text{ is better than } e_2 \\
0, & \text{if } e_2 \text{ is better than } e_1 
\end{cases}
\end{equation}
In each round, we select a reference essay \( r_i \) and pair it with the target essay \( x \) to form the pair \( (x, r_i) \). By swapping the order of the two essays in the prompt, we obtain another pair \( (r_i, x) \). This process yields four comparison results:
\begin{equation}
\begin{cases}
o_1 = C(x, r_1) \\
o_2 = C(r_1, x) \\
o_3 = C(x, r_2) \\
o_4 = C(r_2, x)
\end{cases}
\end{equation}

Define the statistics:
\begin{equation}
\begin{cases}
S_{x>r_i} = o_1 + o_3 \\
S_{r_i>x} = o_2 + o_4 
\end{cases}
\end{equation}
where $S_{x>r_i}$ represents the number of times $x$ is better than $r_i$, and $S_{r_i>x}$ represents the number of times $r_i$ is better than $x$. The final result is defined as:
\begin{equation}
\text{result}(x, r_i) = 
\begin{cases}
r_i \text{>} x, & S_{r_i>x} = 2 \land S_{x>r_i} < 2 \\
r_i \text{>} x, & S_{x>r_i} = 2 \land S_{r_i>x} < 2 \\
r_i \text{=} x, & \text{others}
\end{cases}
\end{equation}

\subsection{Essay Scoring}
We embed the candidate score set information into the data for scoring, fine-tuning the Scorer to endow the model with scoring capabilities. The instruction used for fine-tuning and evaluation in ASAP is as Figure \ref{fig:score_instruction}. And instruction used in HSK is shown in Appendix \ref{sec:HSK}.
\begin{figure}[t]
    \centering
    \includegraphics[width=0.8\columnwidth,height=0.2\textheight]{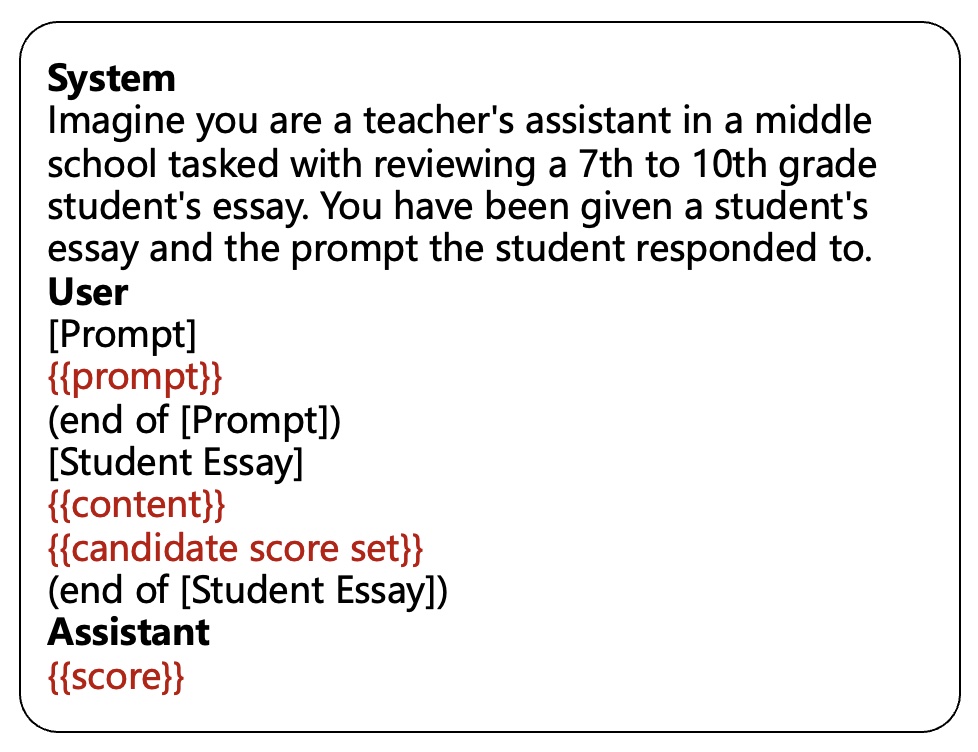}
    \caption{Instruction for fine-tuning the Scorer. Contents to be filled are highlighted in \textcolor{darkred}{red}. }
    \label{fig:score_instruction}
\end{figure}

It is necessary to clarify that due to the overlap of the training data for the Scorer and the Ranker, theoretically, the accuracy of the candidate score set for fine-tuning of the training data is 100\%. Consequently, we introduce some adjustment to the set to lower its accuracy to some extent, thereby achieving better effects.  

\section{Experimental Setup}
\subsection{LLMs}
We conduct experiments using mainstream open-source LLMs for both Chinese and English tasks. For the Scorer, we select \textbf{Qwen2-7B-Instruct} \cite{yang2024qwen} for the Chinese essay scoring task, and select two models of different sizes for English essay scoring: \textbf{LlaMA3.1-8B-Instruct} \cite{grattafiori2024llama} and \textbf{Mistral-NeMo-Instruct-2407} \cite{mistral_nemo_instruct_2407} to demonstrate the general applicability of our method. For the Ranker model, we select \textbf{Qwen2.5-1.5B-Instruct} \cite{yang2024qwen} for Chinese pairwise ranking task, and select \textbf{LlaMA3.2-3B-Instruct} \cite{grattafiori2024llama} for English pairwise ranking task. 

\subsection{Datasets}
We conduct experiments on both Chinese and English datasets. 

For the Chinese data, we utilize the \textbf{HSK} (Hanyu Shuiping Kaoshi) dataset for the Chinese essay scoring task. The HSK dataset originates from the work of \cite{yong2022}, which comprises essay corpora collected from foreign candidates who took the advanced Chinese HSK examination between 1992 and 2005. After cleaning the flag for syntax errors in essays and removing essays with a score of 0 and those with insufficient word counts, we obtain a total of 10,329 essays. Finally, we select the 11 prompts with the largest number of essays for our experiments which contain 8,597 essays. 

The \textbf{ASAP} (Automated Student Assessment Prize) dataset \cite{asap-aes} is famous in the field of English AES, which includes 12,978 essays written by students in grades 7 through 10. These essays are composed in response to 8 prompts covering a variety of genres and score ranges. 
% The statistics of the two datasets are shown in Table \ref{tab:dataset_stats}.
More descriptions of the datasets are provided in Appendix \ref{sec:hsk_data}. 

\subsection{Evaluation Metric}
We use \textbf{Quadratic Weighted Kappa (QWK)} to evaluate the discrepancy between predicted scores and gold scores. This metric is widely adopted in AES tasks for both Chinese and English essays \cite{taghipour2016neural,ridley2020prompt,he2022automated,li2022unified}. 

When generating the candidate score set, we also use \textbf{accuracy} to evaluate the quality of the prediction of our "Binary Search Tree" approach. 

\subsection{Implementation Details}
\textbf{Reference Essay Selecting} Following the previously specified rules, the selected reference scores are shown in Table~\ref{tab:prompt_scores}. For each score, we randomly select \textbf{2 essays} as the reference essays for the current prompt. 
\\
\textbf{Feature Extracting} For both the HSK and ASAP datasets, we select the top 10 final features using LibSVM and F-score, which is shown in Appendix \ref{sec:appendixA}. 
\begin{table}[h]
  \centering
  \small
  \begin{tabular}{ccc}
    \hline
    \textbf{Prompt} & \textbf{Range} & \textbf{Reference Score} \\ \hline
    HSK             & 40-100         & 50,60,70,80,90           \\ 
    ASAP1           & 2-12           & 5,9                      \\ 
    ASAP2           & 1-6            & 3,4                      \\ 
    ASAP3           & 0-3            & 1,2                      \\ 
    ASAP4           & 0-3            & 1,2                      \\ 
    ASAP5           & 0-4            & 2                        \\ 
    ASAP6           & 0-4            & 2                        \\ 
    ASAP7           & 0-30           & 5,10,15,20               \\ 
    ASAP8           & 0-60           & 10,20,30,40,50           \\ \hline
  \end{tabular}
  \caption{The score ranges and corresponding reference scores for both Chinese and English prompts are provided, where ASAP1-ASAP8 represent the prompt IDs in the ASAP dataset.}
  \label{tab:prompt_scores}
\end{table}
\\
\textbf{Pairwise Ranking Data} We set $k=5$ to generate pairwise data for fine-tuning the Ranker, resulting in a rank training dataset that is five times larger than the original scoring training dataset. 
\\
\textbf{Scorer Performance through Candidate Score Calibration } We reduce the accuracy of the Scorer training set according to the accuracy of Ranker's test set. We finally choose to reduce the accuracy of the Scorer training set by 15\% compared to Ranker's test set. 

\begin{table*}[t]
  \centering
  \fontsize{9}{11}\selectfont
  \begin{tabularx}{\textwidth}{Xccccccccccccc}
    \hline
    \textbf{Method} & \textbf{1} & \textbf{2} & \textbf{3} & \textbf{4} & \textbf{5} & \textbf{6} & \textbf{7} & \textbf{8} & \textbf{9} & \textbf{10} & \textbf{11} & \textbf{Avg} \\ \hline
    NPCR & 0.435 & 0.650 & 0.541 & 0.574 & 0.657 & 0.586 & 0.501 & 0.529 & 0.609 & 0.630 & 0.540 & 0.568 \\
    PAES & 0.450 & \textbf{0.730} & 0.690 & 0.679 & 0.658 & 0.663 & 0.751 & 0.662 & 0.702 & 0.720 & 0.458 & 0.651 \\
    Vanilla & 0.625 & 0.725 & 0.774 & 0.696 & 0.812 & 0.788 & 0.854 & 0.758 & 0.754 & 0.776 & 0.643 & 0.746 \\ \hline
    \textbf{RTS} & \textbf{0.657} & 0.716 & \textbf{0.796} & \textbf{0.706} & \textbf{0.823} & \textbf{0.789} & \textbf{0.863} & \textbf{0.797} & \textbf{0.755} & \textbf{0.779} & \textbf{0.732} & \textbf{0.765} \\ \hline
  \end{tabularx}
  \caption{Performance of our method on the HSK dataset. The bolded data are the best performing results among all Models.The scorer model used by RTS is \textbf{Qwen2-7B-Instruct}.}
  \label{tab:hsk}
\end{table*}
\begin{table*}[t]
  \centering
  \small
  \fontsize{9}{10}\selectfont
  \begin{tabularx}{\textwidth}{Xlccccccccc}
    \hline
    \small
    \textbf{Model} & \textbf{Method} & \textbf{1} & \textbf{2} & \textbf{3} & \textbf{4} & \textbf{5} & \textbf{6} & \textbf{7} & \textbf{8} & 
    \textbf{Avg} \\ \hline
    R$^{2}$BERT & - & 0.817 & 0.719 & 0.698 & 0.845 & 0.841 & 0.847 & 0.839 & 0.744 & 0.794 \\
    NPCR & - & 0.856 & 0.750 & 0.756 & 0.851 & 0.847 & 0.858 & 0.838 & 0.779 & 0.817 \\ \hline
    \multirow{2}{*}{LlaMA3.1-8B-Instruct} & Vanilla & 0.822 & 0.688 & 0.724 & 0.826 & 0.806 & 0.845 & 0.830 & 0.706 & 0.781 \\
    & \textbf{RTS} & \textbf{0.840} & \textbf{0.712} & \textbf{0.752} & \textbf{0.844} & \textbf{0.831} & \textbf{0.848} & 0.830 & \textbf{0.732} & \textbf{0.798} \\ \hline
    \multirow{2}{*}{Mistral-NeMo-Instruct-2407} & Vanilla & 0.823 & 0.688 & 0.705 & 0.836 & 0.801 & 0.838 & \textbf{0.841} & 0.728 & 0.783 \\
    & \textbf{RTS} & \textbf{0.835} & \textbf{0.710} & \textbf{0.730} & \textbf{0.840} & \textbf{0.821} & \textbf{0.839} & 0.838 & \textbf{0.740} & \textbf{0.794} \\
    \hline
  \end{tabularx}
  \caption{Performance of our method on the ASAP dataset. The first row shows the NPCR method which is SOTA method on small models. the bolded data are the results where our method significantly outperforms Vanilla's method. The two LLMs used here are both their Instruct versions.}
  \label{tab:asap}
\end{table*}

\subsection{Comparing Methods}
We compare RTS with other excellent supervised method. 

\textbf{R${^{2}}$BERT} \cite{yang2020enhancing} Significant improvements are achieved by modifying the scoring loss to a combination of pairwise ranking and scoring losses.

\textbf{NPCR} \cite{xie2022automated} This is the state-of-the-art supervised prompt-specific AES method in the ASAP dataset. They utilized up to 50 reference essays to compare with the target essay, achieving excellent results. We also apply this method to the HSK dataset to compare its performance. 

\textbf{PAES} \cite{ridley2020prompt} A highly effective cross-prompt AES method that also incorporates features. The features we used for the ASAP dataset are adapted from this work. 

\textbf{Vanilla} Fine-tuning the model directly without incorporating the candidate score set. 

% \textbf{SFT with Features} Based on Vanilla SFT, we added the corresponding Features of articles in Instruction to fine-tune the model. 

\section{Results and Analysis}

\subsection{Main Results}
The final experimental results are shown in Table \ref{tab:hsk} and Table \ref{tab:asap}. Overall, RTS is able to outperform the Vanilla method both in average QWK and QWK on almost all prompts, which shows that RTS has the enhancement effect not only on different datasets in different languages, but also on different LLMs.

Expanding on this, the improvement of HSK on average QWK is 1.9\% ${(74.6\% \to 76.5\%)}$, and the improvement of ASAP is 1.7\% ${(78.1\% \to 79.8\%)}$ and 1.1\% ${(78.3\% \to 79.4\%)}$ respectively, note that compared to the Vanilla method, RTS's improvement in average QWK is similar across datasets and models. In terms of each dataset, in HSK, RTS boosts ranged from 0.9\% to 8.9\%, with prompt 11 boosting the most by 8.9\% $(64.3\% \to 73.2\%)$. In the ASAP dataset, the boost ranges from 1.4\% to 2.8\%, with the largest boost being 2.8\% $(72.4\% \to 75.2\%)$ for LlaMA3.1-8B-Instruct on prompt3. All of these improvements indicate that the improvement effect of RTS is similar across different data and has good cross-language capabilities. 
% However, it is interesting to note that the improvement effect of Mistral-Nemo is less than that of LlaMA3.1-8B and Qwen2-7B with fewer parameters, suggesting that RTS may be more suitable for LLMs with fewer parameters

It is also worth noting that, on the HSK dataset, the RTS method also significantly outperforms the results of all the small models that perform extremely well on the ASAP dataset $(56.8\%,65.1\% \to 74.6\%,76.5\%)$, which demonstrates that LLM has a very high potential for Chinese AES tasks. 

However, if we look closely, we can see that except for prompt 6, 9 and 10 in HSK and prompt 7 in ASAP, where there is almost no improvement ${(78.8\%\to78.9\%)}$, prompt 2 decreases by 0.9\% compared to Vanilla ${(72.5\%\to71.6\%)}$. We will analyze the reasons for this phenomenon in the next section with another set of experiments.

%\subsection{Analysis}
%By designing other experiments, we have analyzed some of the reasons why RTS has achieved such an improvement.
%\subsubsection{Ceiling Analysis}
\subsection{Upper Bound Analysis}
\begin{figure}[h]
    \centering
    \includegraphics[width=\columnwidth]{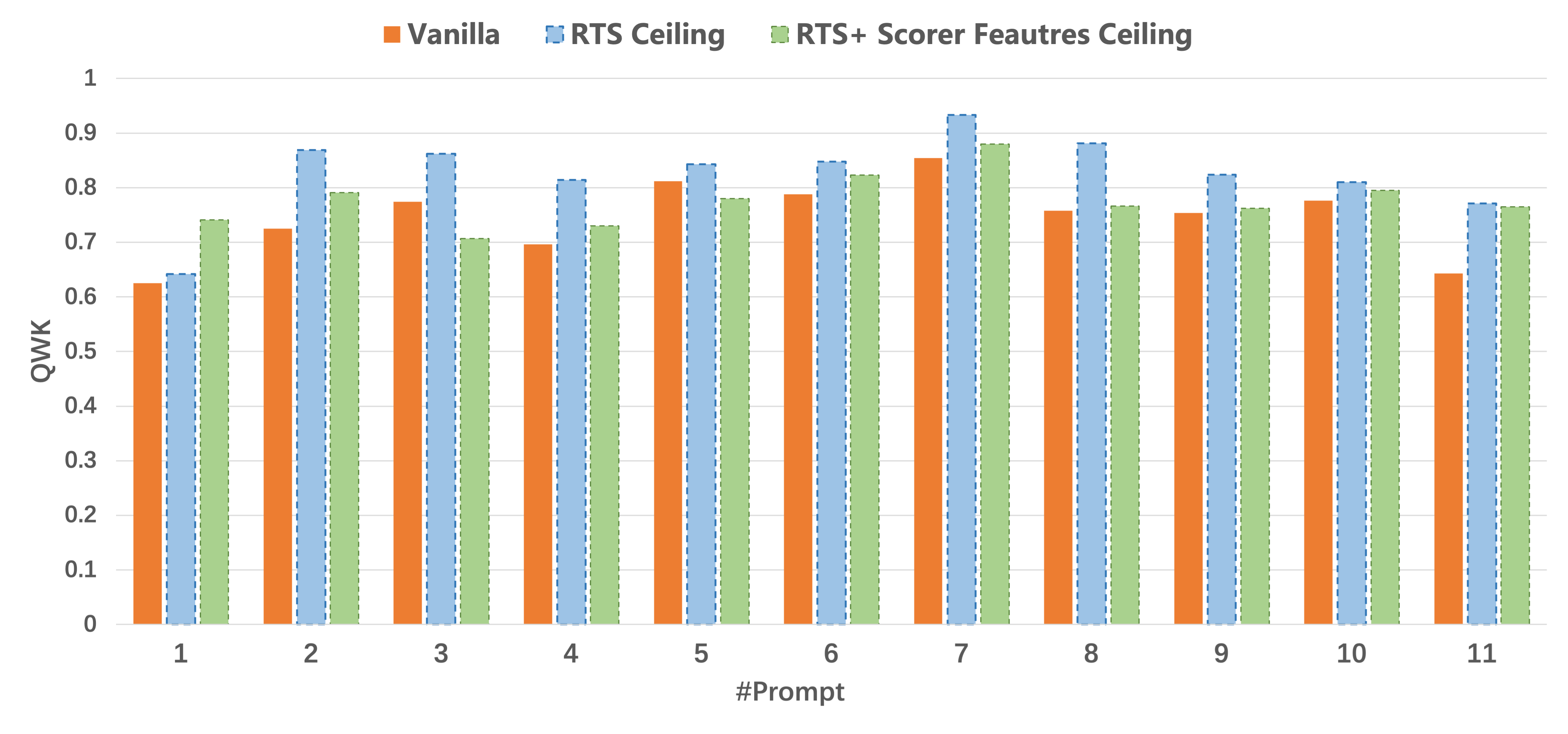}
    \caption{The results in the HSK dataset of two ceiling experiments are presented.}
    \label{fig:ceiling}
\end{figure}

Before verifying the RTS method, we first fine-tune Scorer with a candidate score set with 100\% accuracy in order to see if our hypothesis is reasonable. We also validate the model by adding features to the Scorer in order to determine whether features are applicable in the RTS to where they are added to the Scorer. The result in the HSK dataset is shown in Figure \ref{fig:ceiling}. 

As shown in the figure, all prompts RTS have extremely high ceilings, so RTS methods are probably viable. However, because the accuracy of the candidate score set cannot reach the ideal state, it is difficult to reach the ceiling in practice. On the other hand, adding features to the Scorer downs the ceiling of the RTS method, which explains why we do not add features to the Scorer in the final RTS method.

\subsection{Ablation Study of Features}
%In addition to the second ceiling analysis presented in \textbf{5.2.1}, 
Further, we explore the effects of incorporating features into different components of RTS, as shown in Table \ref{tab:features_scorer}. 
\begin{table}[htbp]
  \centering
  \small
  % \fontsize{9}{13}\selectfont
  \begin{tabularx}{\columnwidth}{Xc}
    \hline
    \textbf{Method} & \textbf{Avg} \\ \hline
    Vanilla & 0.746 \\ 
    RTS & 0.765 \\ 
    \phantom{11} + Scorer Features & 0.719 \\
    \phantom{11} - Ranker Features & 0.751 \\ \hline
  \end{tabularx}
  \caption{Results of adding features to scorer and removing features from ranker on HSK.}
  \label{tab:features_scorer}
\end{table}

\begin{figure}[h]
    \centering
    \includegraphics[width=0.9\columnwidth,height=0.2\textheight]{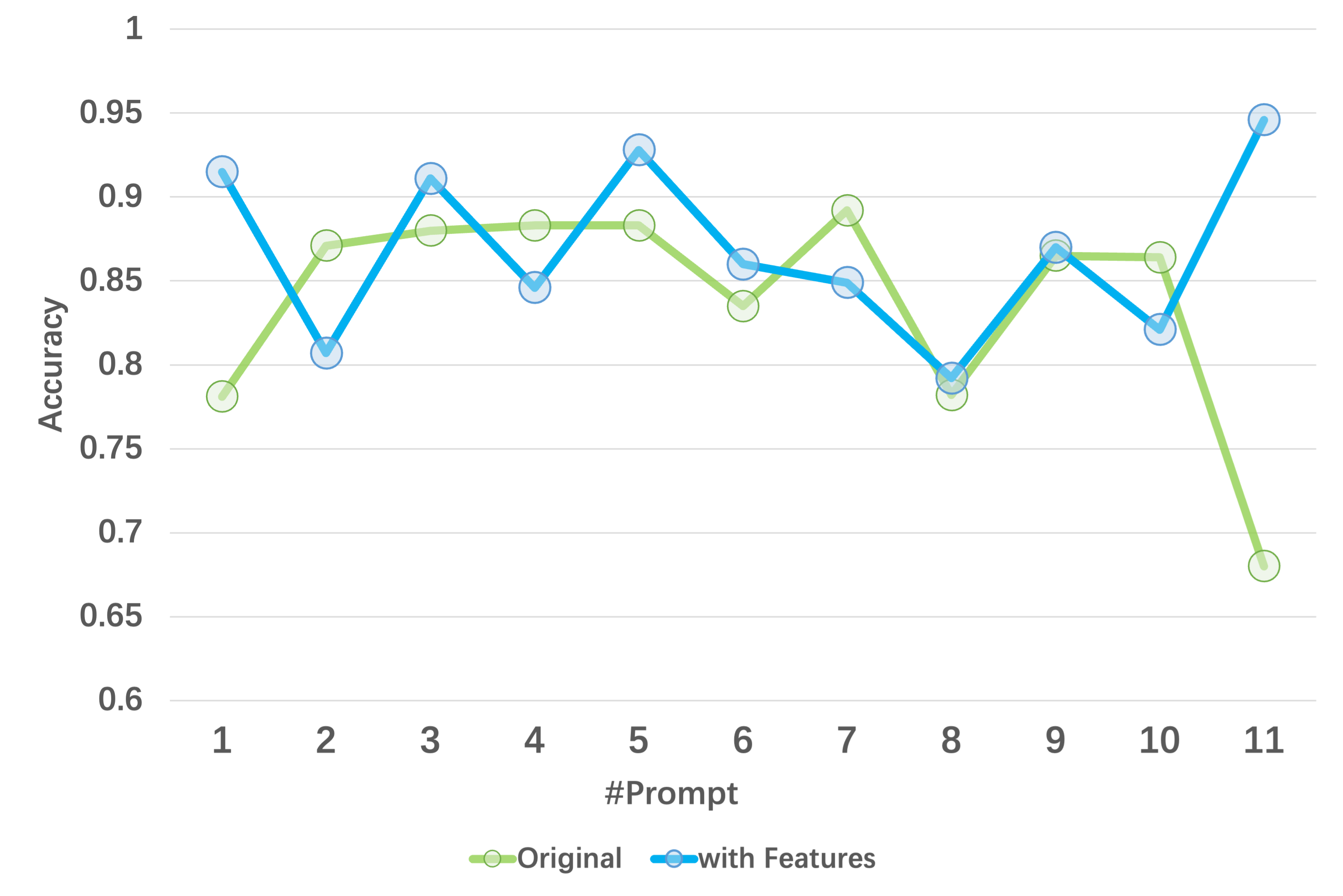}
    \caption{The impact of adding features on the Ranker's classification accuracy. The average accuracy rate after adding Features is 3.1\% higher than that without adding features (${83.7\% \to 86.8\%}$).}
    \label{fig:features_rank}
\end{figure}

As can be seen in Table \ref{tab:features_scorer}, RTS decreases by 1.4\% $(76.5\% \to 75.1\%)$ after removing Ranker features. A feasible approach to addressing this issue is to determine the accuracy of pairwise rankings with and without features. The result of this experiment on HSK is shown in Figure \ref{fig:features_rank}. 

From the perspective of average accuracy, Ranker's ranking ability is significantly improved after the addition of features, especially on some prompts. However, we can also clearly observe that the accuracy even decreases on four prompts, with prompt 2 decreasing the most significantly $({87.1\% \to 80.7\%)}$. Not only does this shows that features is not facilitated in some prompt, but it also explains why RTS performance drops on prompt 2 in HSK dataset, which is observed in \textbf{5.1}.

\subsection{Scorer Performance through Candidate Score Calibration} 
\begin{figure}[h]
    \centering
    \includegraphics[width=0.9\columnwidth,height=0.25\textwidth]{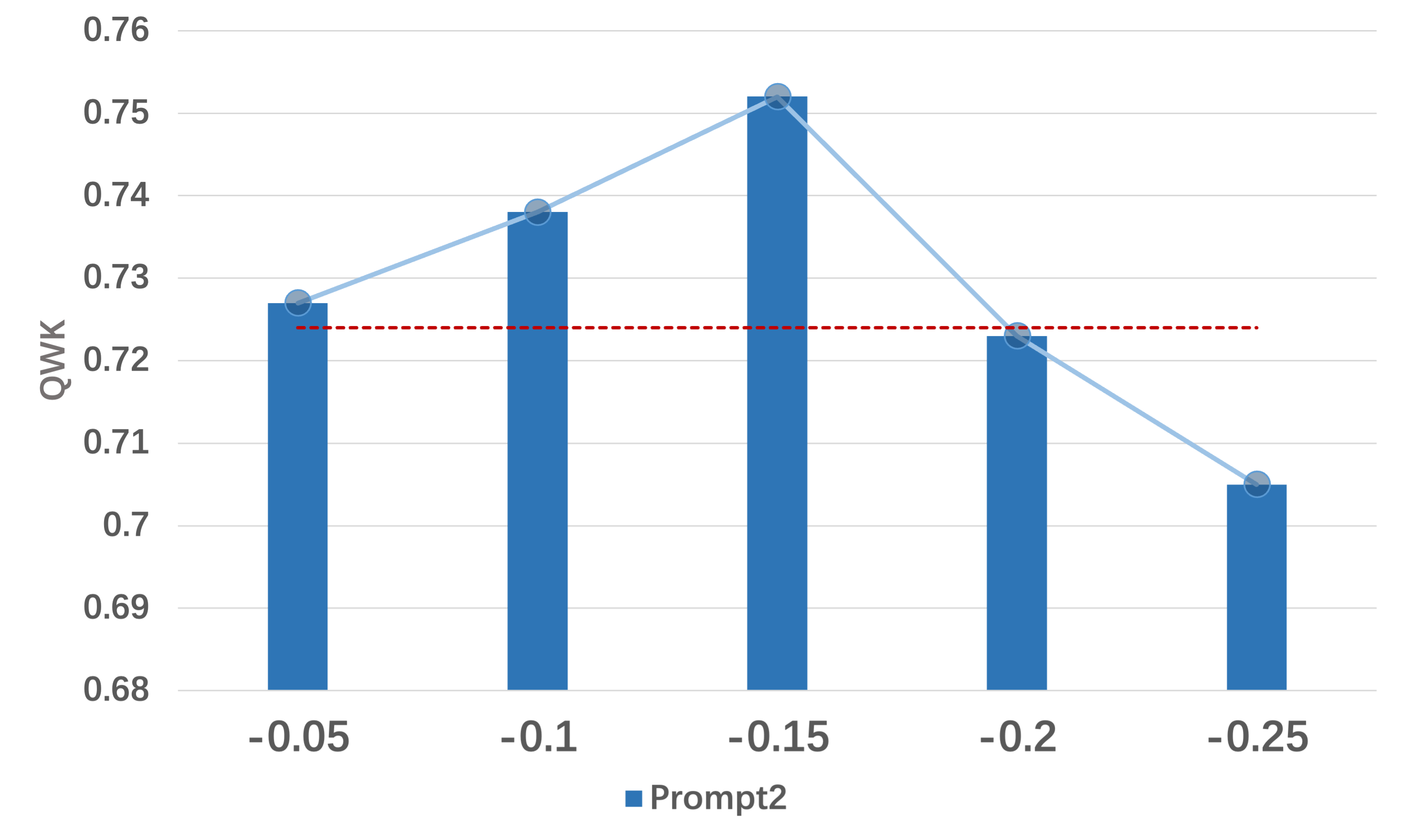}
\caption{This is the result on Prompt 2 of the ASAP dataset. The x-axis represents the degree of adjustment. For example, 
"$-0.1$" indicates that the accuracy of the Ranker's test data is $a$, while the accuracy of the Scorer's candidate score set is "$a-0.1$".}
\label{fig:gap}
\end{figure}
In the process of fine-tuning the Scorer, we observe that adjusting the accuracy of the candidate score set used in some of the fine-tuning data based on the accuracy of the Ranker, is able to improve the results of the Scorer. The results from the experiment, with different adjustment values, are shown in Figure \ref{fig:gap}. As illustrated, the Scorer's performance is optimal when the accuracy of the Ranker's test data differs by 0.15 from that of the Scorer's candidate score set.

\subsection{Other Ranking and Scoring Combined Methods}
We explore some other methods that can combine the ranking task and the scoring task on LLMs \cite{yang2020enhancing,xie2022automated}:

\textbf{Scoring In Multiple Essays} When we assume that the model can automatically learn the order among multiple essays from the history, we give the model 5 different essays at a time and let the model score them.

\textbf{Simultaneous Generation of Scores And Rankings} Based on the above assumption, we propose another method: generate both scores and rankings in all previous essays.

\textbf{Both In One LLM} Starting from the idea of Multi-Task, it is easy to think of a way to fine-tune the Scorer with pairwise ranking data. Therefore, we divide this method into two types of fine-tuning phases: first ranking and then scoring, and first scoring and then ranking.

The results of the above method compared to the RTS method on the HSK dataset are shown in Table \ref{tab:diff_method}. It can be clearly seen that the first two results prove that the assumption mentioned above is not true, and both methods have a lowering effect on the model. For the latter two methods, the S1R2 method has a better improvement, but it is still only 0.6\% ${(74.6\% \to 75.2\%)}$ much less than the 1.9\% ${(74.6\% \to 76.5\%)}$ boost of the RTS. The above results illustrate that of all the methods combining ranking and scoring, RTS is the one that performs best on LLMs.
\begin{table}[htbp]
  \centering
  \small
  % \fontsize{9}{13}\selectfont
  \begin{tabularx}{\columnwidth}{Xc}
    \hline
    \textbf{Method} & \textbf{Avg} \\ \hline
    Vanilla & 0.746 \\ \hline
    Scoring in 5 Essays & 0.656 \\
    Simultaneous Generation & 0.676 \\ 
    R1S2 & 0.509 \\
    S1R2 & 0.752 \\ \hline
    RTS & 0.765 \\ \hline
  \end{tabularx}
  \caption{Results of other methods on the HSK dataset. \textbf{R1S2} indicates ranking first, then scoring. \textbf{S1R2} indicates scoring first, then ranking}
  \label{tab:diff_method}
\end{table}

% \subsubsection{Cost Analysis}
% In the NPCR method, the authors used 50 reference essays to achieve the SOTA result of 0.817. In our method, on the other hand, we achieved 0.798 using a maximum of 10 and a minimum of 2 reference essay, very close to the NPCR method when 50 essays are used.

\section{Conclusion}
This study introduces RTS (Rank-Then-Score), a novel LLM-based fine-tuning method for Automated Essay Scoring (AES) across Chinese and English datasets. RTS combines two specialized LLMs: one fine-tuned for essay ranking and another fine-tuned for scoring, achieving superior improvements. Experiments show RTS significantly outperforms traditional Vanilla fine-tuning, particularly in Chinese dataset. Key findings include: (1) RTS has the best AES performance on LLMs; (2) Integrating features into the Ranker enhances quality discrimination more effectively than adding them to the Scorer; (3) RTS surpasses other ranking-scoring combinations on LLMs by enabling seamless integration with human grading standards. The method demonstrates exceptional cross-lingual adaptability and precision, offering a robust solution for scenarios requiring nuanced essay evaluation. This dual-model approach addresses subtle quality distinctions while maintaining alignment with manual assessment practices, marking a notable advancement in AES technology on LLMs.

\section*{Limitations}

Firstly, our architecture comprises two LLMs. Although the Ranker employs a relatively smaller model, there is still room for optimization in the size of both models. Encouragingly, we experiment with using Qwen2.5-1.5B-Instruct as the Scorer, and on the HSK dataset, the Vanilla method still achieves an average QWK of 0.741. This demonstrates that our approach has the potential to perform well on even smaller LLMs. Such models can be more effectively utilized in practical applications.

Another issue that requires attention is the selection of reference essays. Although we achieve satisfactory results by randomly selecting reference essays, it is still necessary to explore whether different methods of selecting reference essays will significantly impact our approach.

\section*{Ethics Statement}

\textbf{Potential Risks} Our method cannot guarantee fair evaluation, meaning that RTS may reinforce the LLMs' tendency to favor certain social groups in scoring. For example, the predicted results may assign higher scores to groups with specific L1 (first language) backgrounds compared to other groups. Additionally, the datasets we used (ASAP and HSK) may disproportionately represent certain demographic groups, potentially leading to biased conclusions.
\\
\textbf{Use of Scientific Artifact} We utilize the open-source scikit-learn package \cite{Pedregosa2011ScikitlearnML} to compute the Quadratic Weighted Kappa (QWK). For our experiments, we employ the ASAP dataset \cite{asap-aes} and the HSK dataset \cite{yong2022}, both of which are available for non-commercial research purposes. Both ASAP and HSK replace personally identifiable information in the essays with symbols. Features used in ASAP and the types of features referenced in HSK both originate from open-source code \cite{ridley2020prompt,li2022unified}. The large language models used in this study, LlaMA 3 \cite{grattafiori2024llama}, Mistral \cite{mistral_nemo_instruct_2407}, and Qwen2 \cite{yang2024qwen}, are licensed under the LlaMA 3 Community license and Apache-2.0 license, respectively. Alllicenses permit their use for research purposes.
\\
\textbf{Computational Budget} We utilize two NVIDIA A40 GPUs for model fine-tuning and a single NVIDIA A40 GPU for inference of each model, including Qwen2.5-1.5B-Instruct,Qwen2-7B-Instruct, LlaMA3.2-3B-Instruct, LlaMA3.1-8B-Instruct, and Mistral-NeMo-Instruct-2407. Each batch contains 8 samples. Fine-tuning the RTS method on both datasets take approximately 2 hours, while inference, including both the Ranker and Scorer, take a maximum of 12 seconds per sample. However, the inference time may vary depending on the model architecture and acceleration methods employed.

\bibliography{custom}
\clearpage

\appendix

\newpage
\section{Types of Features}
\label{sec:appendixA}
\begin{table}[h!]
  \centering
  \fontsize{10}{12}\selectfont
  \begin{tabularx}{\columnwidth}{clX}
    \hline
    \textbf{Idx} & \textbf{Dim} & \textbf{Feature Description} \\ \hline
    1 & 1 & Total number of characters \\
    2 & 1 & Number of character types \\
    3 & 1 & Type Token Ratio (TTR) \\
    4 & 1 & Average number of strokes \\
    5 & 1 & Weighted average number of strokes \\
    6 & 25 & Number of characters with different strokes \\
    7 & 25 & Proportion of characters with different strokes \\
    8 & 1 & Average character frequency \\
    9 & 1 & Weighted average character frequency \\
    10 & 1 & Number of single characters \\
    11 & 1 & Proportion of single characters \\
    12 & 1 & Number of common characters \\
    13 & 1 & Proportion of common characters \\
    14 & 1 & Number of unregistered characters \\
    15 & 1 & Proportion of unregistered characters \\
    16 & 1 & Number of first-level characters \\
    17 & 1 & Proportion of first-level characters \\
    18 & 1 & Number of second-level characters \\
    19 & 1 & Proportion of second-level characters \\
    20 & 1 & Number of third-level characters \\
    21 & 1 & Proportion of third-level characters \\
    22 & 1 & Number of fourth-level characters \\
    23 & 1 & Proportion of fourth-level characters \\
    24 & 1 & Average character level \\
    \hline
  \end{tabularx}
  \caption{Character features description.}
  \label{tab:char_features}
  % \begin{tabularx}{\columnwidth}{clX}
  %   \\
  %   \hline
  %   \textbf{Idx} & \textbf{Dim} & \textbf{Feature Description} \\ \hline
  %   1 & 1 & Total number of sentences \\  
  %   2 & 1 & Average characters in a sentence \\  
  %   3 & 1 & Average words in a sentence \\  
  %   4 & 1 & Maximum characters in a sentence \\  
  %   5 & 1 & Maximum words in a sentence \\  
  %   6 & 1 & Number of clauses \\  
  %   7 & 1 & Average characters in a clause \\  
  %   8 & 1 & Average words in a clause \\  
  %   9 & 1 & Maximum characters in a clause \\  
  %   10 & 1 & Maximum words in a clause \\  
  %   11 & 30 & Sentence length distribution \\  
  %   12 & 1 & Average syntax tree height \\  
  %   13 & 1 & Maximum syntax tree height \\  
  %   14 & 1 & Syntax tree height $\leq$ 5 ratio \\  
  %   15 & 1 & Syntax tree height $\leq$ 10 ratio \\  
  %   16 & 1 & Syntax tree height $\leq$ 15 ratio \\  
  %   17 & 1 & Syntax tree height $\geq$ 16 ratio \\  
  %   18 & 14 & Dependency distribution \\
  %   \hline
  % \end{tabularx}
  % \caption{Sentence features description.}
  % \label{tab:char_features}
  % \begin{tabularx}{\columnwidth}{clX}
  %   \\
  %   \hline
  %   \textbf{Idx} & \textbf{Dim} & \textbf{Feature Description} \\ \hline
  %   1 & 1 & Total number of paragraphs \\ 
  %   2 & 1 & Average characters in a paragraph \\ 
  %   3 & 1 & Average words in a paragraph \\ 
  %   4 & 1 & Maximum characters in a paragraph \\ 
  %   5 & 1 & Maximum words in a paragraph \\ \hline
  % \end{tabularx}
  % \caption{Paragraph features description.}
  % \label{tab:char_features}
\end{table}
\begin{table}[b!]
  \centering
  \fontsize{10}{12}\selectfont
  \begin{tabular}{cll}
    \hline
    \textbf{Idx} & \textbf{Dim} & \textbf{Feature Description} \\ \hline
    1 & 1 & Total number of sentences \\  
    2 & 1 & Average characters in a sentence \\  
    3 & 1 & Average words in a sentence \\  
    4 & 1 & Maximum characters in a sentence \\  
    5 & 1 & Maximum words in a sentence \\  
    6 & 1 & Number of clauses \\  
    7 & 1 & Average characters in a clause \\  
    8 & 1 & Average words in a clause \\  
    9 & 1 & Maximum characters in a clause \\  
    10 & 1 & Maximum words in a clause \\  
    11 & 30 & Sentence length distribution \\  
    12 & 1 & Average syntax tree height \\  
    13 & 1 & Maximum syntax tree height \\  
    14 & 1 & Syntax tree height $\leq$ 5 ratio \\  
    15 & 1 & Syntax tree height $\leq$ 10 ratio \\  
    16 & 1 & Syntax tree height $\leq$ 15 ratio \\  
    17 & 1 & Syntax tree height $\geq$ 16 ratio \\  
    18 & 14 & Dependency distribution \\
    \hline
  \end{tabular}
  \caption{Sentence features description.}
  \label{tab:char_features}
\end{table}
\begin{table}[h!]
  \centering
  \fontsize{10}{12}\selectfont
  \begin{tabular}{cll}
    \hline
    \textbf{Idx} & \textbf{Dim} & \textbf{Feature Description} \\ \hline
    1 & 1 & Total number of words \\
    2 & 1 & Number of word types \\
    3 & 1 & Type Token Ratio (TTR) \\
    4 & 1 & Average word length \\
    5 & 1 & Weighted average word length \\
    6 & 1 & Average word frequency \\
    7 & 1 & Weighted average word frequency \\
    8 & 1 & Number of single-character words \\
    9 & 1 & Proportion of single-character words \\
    10 & 1 & Number of two-character words \\
    11 & 1 & Proportion of two-character words \\
    12 & 1 & Number of three-character words \\
    13 & 1 & Proportion of three-character words \\
    14 & 1 & Number of four-character words \\
    15 & 1 & Proportion of four-character words \\
    16 & 1 & Number of multi-character words \\
    17 & 1 & Proportion of multi-character words \\
    18 & 1 & Number of idioms \\
    19 & 1 & Number of single words \\
    20 & 1 & Proportion of single words \\
    21 & 1 & Number of unregistered words \\
    22 & 1 & Proportion of unregistered words \\
    23 & 1 & Number of first-level words \\
    24 & 1 & Proportion of first-level words \\
    25 & 1 & Number of second-level words \\
    26 & 1 & Proportion of second-level words \\
    27 & 1 & Number of third-level words \\
    28 & 1 & Proportion of third-level words \\
    29 & 1 & Number of fourth-level words \\
    30 & 1 & Proportion of fourth-level words \\
    31 & 1 & Average word level \\
    32 & 57 & Number of words with different POS \\
    33 & 57 & Proportion of words with different POS \\
    \hline
  \end{tabular}
  \caption{Word features description.}
  \label{tab:char
  _features}
\end{table}
\begin{table}[h]
  \centering
  \fontsize{10}{12}\selectfont
  \begin{tabular}{cll}
    \hline
    \textbf{Idx} & \textbf{Dim} & \textbf{Feature Description} \\ \hline
    1 & 1 & Total number of paragraphs \\ 
    2 & 1 & Average characters in a paragraph \\ 
    3 & 1 & Average words in a paragraph \\
    4 & 1 & Maximum characters in a paragraph \\ 
    5 & 1 & Maximum words in a paragraph \\ 
    \hline
  \end{tabular}
  \caption{Paragraph features description.}
  \label{tab:char_features}
\end{table}
\begin{table*}[t]
\centering
\begin{tabular}{ccc}
\hline
\textbf{Idx} & \textbf{Feature Name} & \textbf{Full Name} \\ \hline
1 & mean\_word & Mean Word Length \\  
2 & word\_var & Word Variance \\  
3 & mean\_sent & Mean Sentence Length \\  
4 & sent\_var & Sentence Variance \\  
5 & ess\_char\_len & Essential Character Length \\  
6 & word\_count & Word Count \\  
7 & prep\_comma & Preposition to Comma Ratio \\  
8 & unique\_word & Unique Word Count \\  
9 & clause\_per\_s & Clauses per Sentence \\  
10 & mean\_clause\_l & Mean Clause Length \\  
11 & max\_clause\_in\_s & Maximum Clauses in a Sentence \\  
12 & spelling\_err & Spelling Error Count \\  
13 & sent\_ave\_depth & Sentence Average Depth \\  
14 & ave\_leaf\_depth & Average Leaf Depth \\  
15 & automated\_readability & Automated Readability Index \\  
16 & linsear\_write & Linsear Write Formula \\  
17 & stop\_prop & Stopword Proportion \\  
18 & positive\_sentence\_prop & Positive Sentence Proportion \\  
19 & negative\_sentence\_prop & Negative Sentence Proportion \\  
20 & neutral\_sentence\_prop & Neutral Sentence Proportion \\  
21 & overall\_positivity\_score & Overall Positivity Score \\  
22 & overall\_negativity\_score & Overall Negativity Score \\ \hline
\end{tabular}
\caption{Text Statistical Features and Their Full Names}
\label{tab:features}
\end{table*}
\begin{table*}[h]
  \centering
  \fontsize{11}{13}\selectfont
  \begin{tabularx}{\columnwidth}{cX}
    \hline
    \small
    \textbf{Dataset} & \textbf{Features} \\ \hline
    \multirow{9}*{HSK} & Total Word Count, \\
        & Character TTR (Type-Token Ratio), \\
        & Word TTR, \\
        & Proportion of Advanced Characters, \\
        & Proportion of Advanced Words, \\
        & Character-Level Weighted Score, \\
        & Word-Level Weighted Score, \\
        & Number of Sentences, \\
        & Average Syntactic Tree Height, \\
        & Maximum Syntactic Tree Height. \\
    \hline
    \multirow{9}*{ASAP} & Mean Word Length, \\
        & Mean Sentence Length,\\
        & Essay Character Length, \\
        & Total Word Count, \\
        & Number of Unique Words, \\
        & Clauses per Sentence, \\
        & Spelling Errors, \\
        & Sentence Average Syntactic Depth, \\
        & Automated Readability Index (ARI), \\
        & Linsear Write Formula. \\ \hline
  \end{tabularx}
  \caption{Selected features on two datasets.} %are shown above.}
  \label{tab:filtered_features}
\end{table*}
\clearpage

\newpage
\section{Instructions}
\label{sec:HSK}
The following are the instructions for the Ranker and Scorer in RTS method for HSK, Vanilla method for HSK, Vanilla method for ASAP, and the two methods: Scoring in 5 essays and Simultaneous Generation.

\begin{figure}[h!]
    \centering
    \includegraphics[width=\columnwidth]{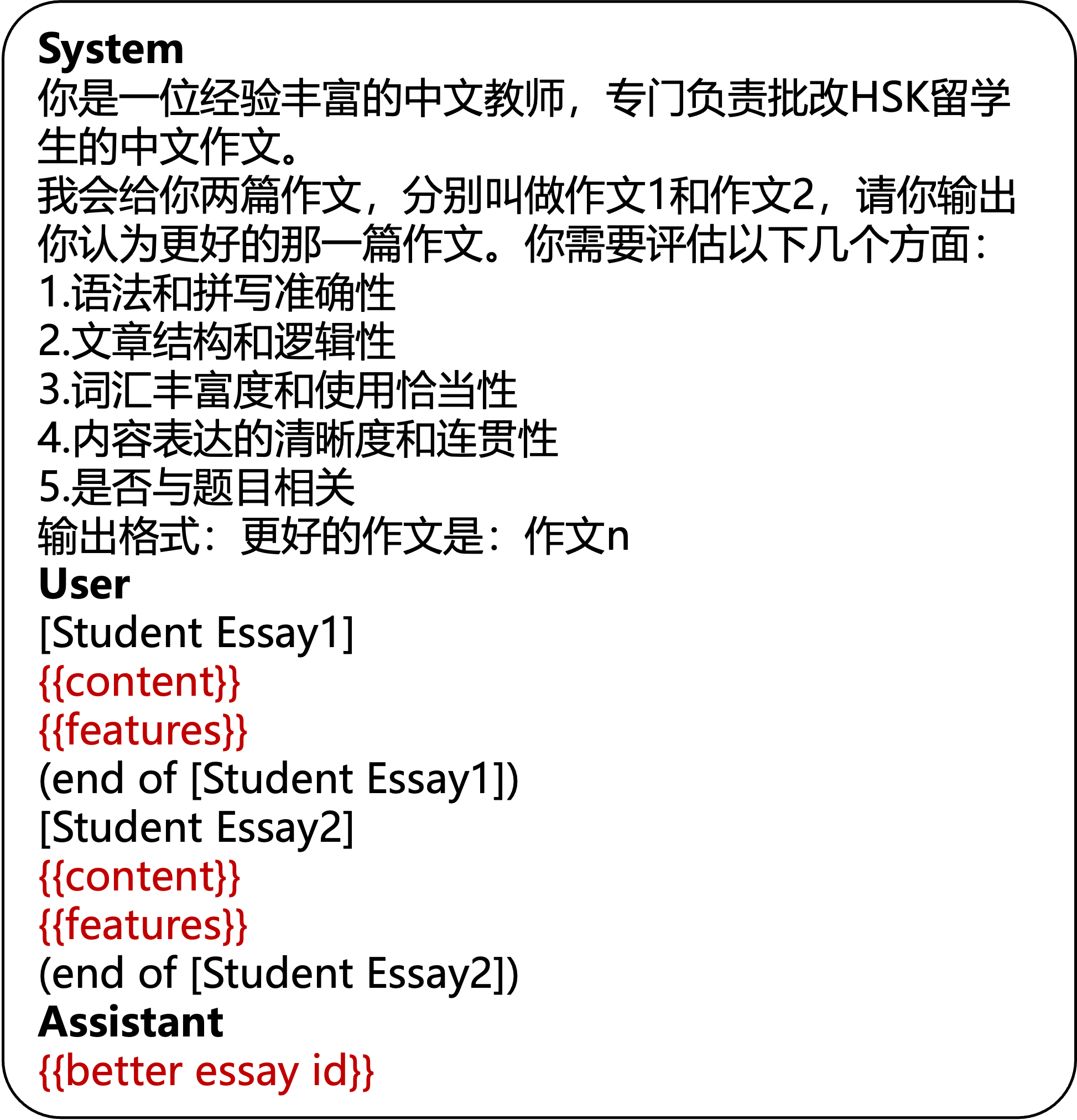}
    \caption{Instruction for the Ranker in RTS for HSK. Contents to be filled are highlighted in \textcolor{darkred}{red}.}
\end{figure}

\begin{figure}[h!]
    \includegraphics[width=\columnwidth]{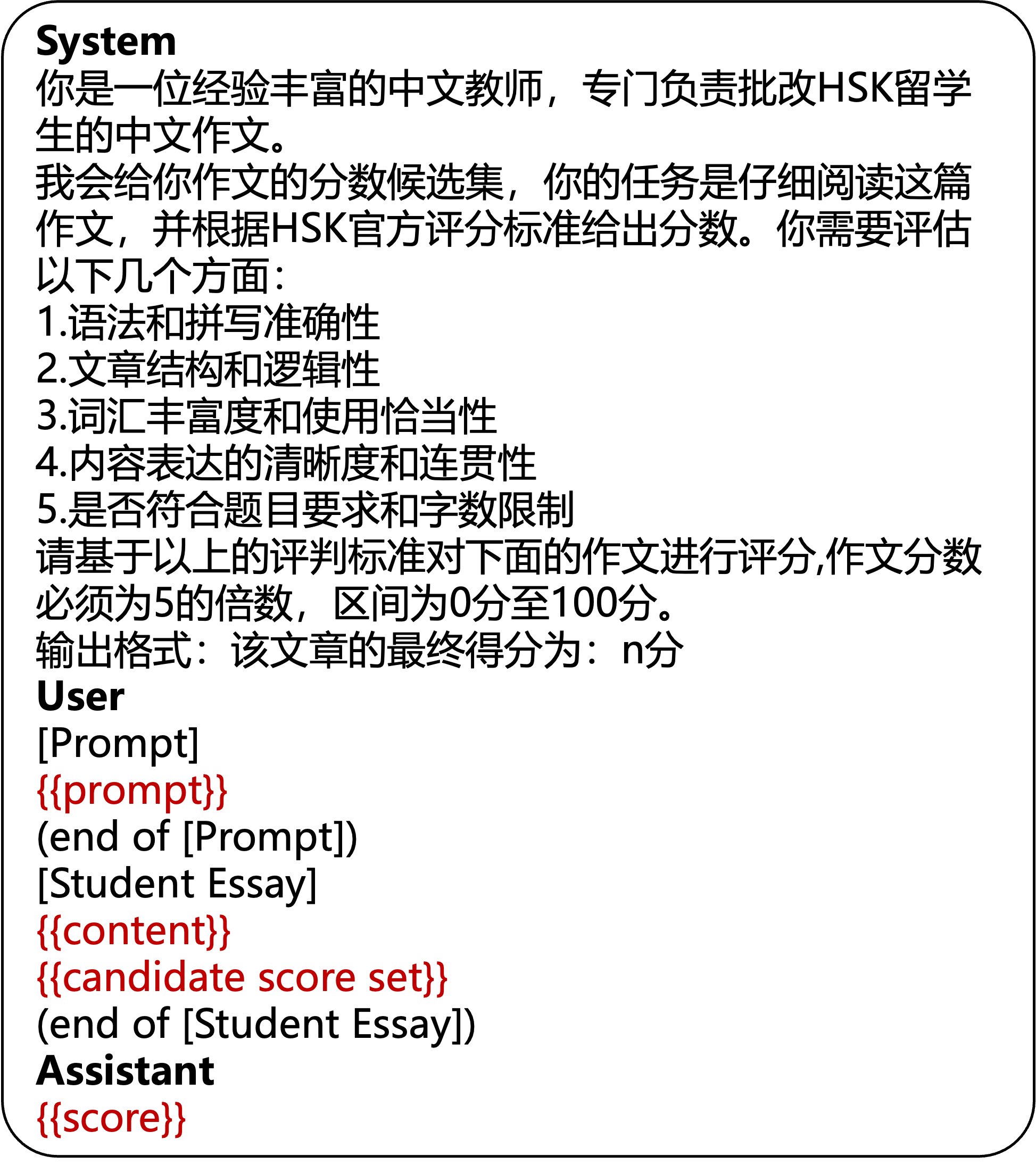}
    \caption{Instruction for the Scorer in RTS for HSK. Contents to be filled are highlighted in \textcolor{darkred}{red}.}
    \label{fig:hsk_ranker}
\end{figure}

\begin{figure}[t]
    \centering
    \includegraphics[width=\columnwidth]{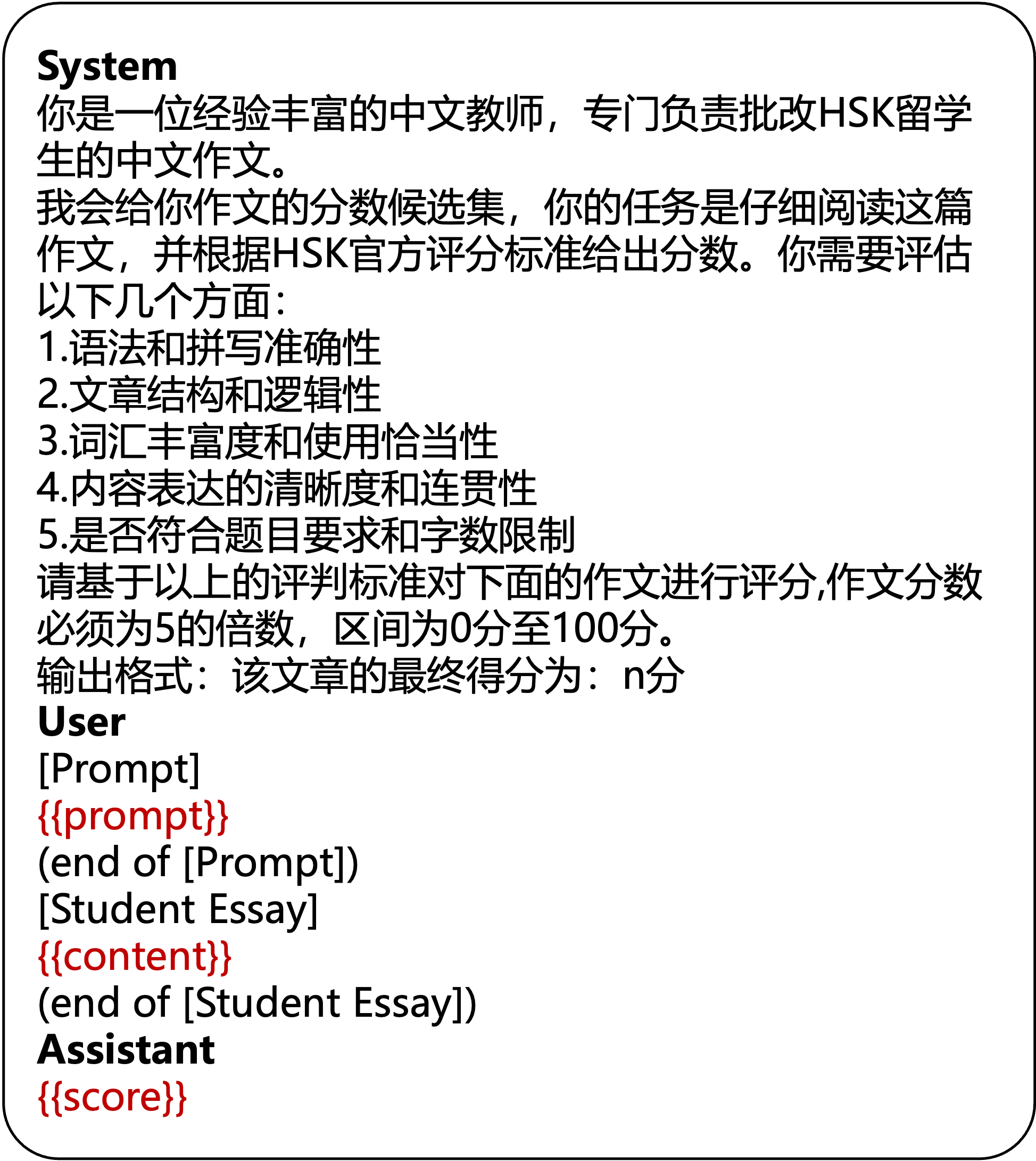}
    \caption{Instruction for Vanilla for HSK. Contents to be filled are highlighted in \textcolor{darkred}{red}.}
\end{figure}
\begin{figure}[h]
    \centering
    \includegraphics[width=\columnwidth]{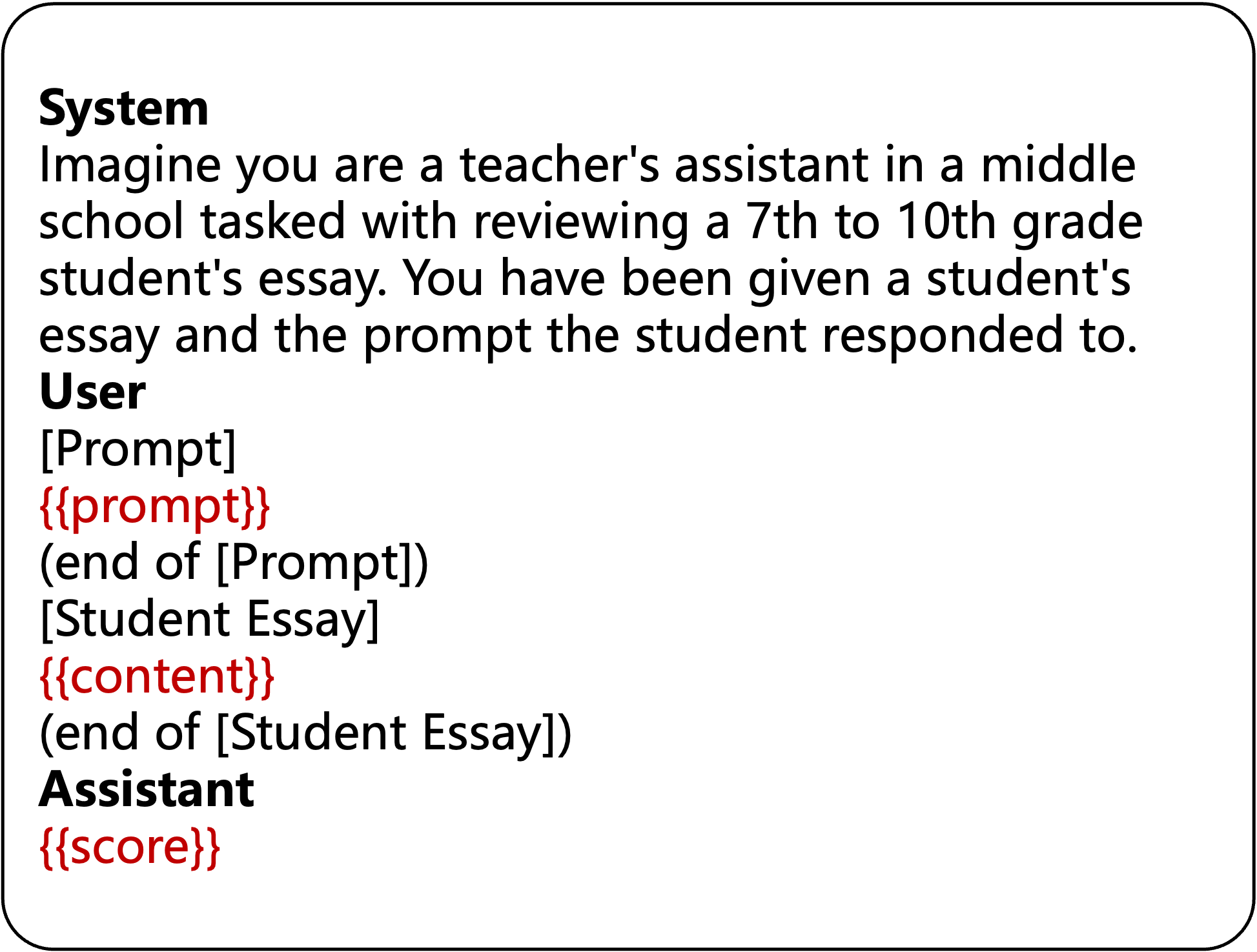}
    \caption{Instruction for Vanilla for ASAP. Contents to be filled are highlighted in \textcolor{darkred}{red}.}
\end{figure}

\begin{figure}[t]
    \centering
    \includegraphics[width=\columnwidth]{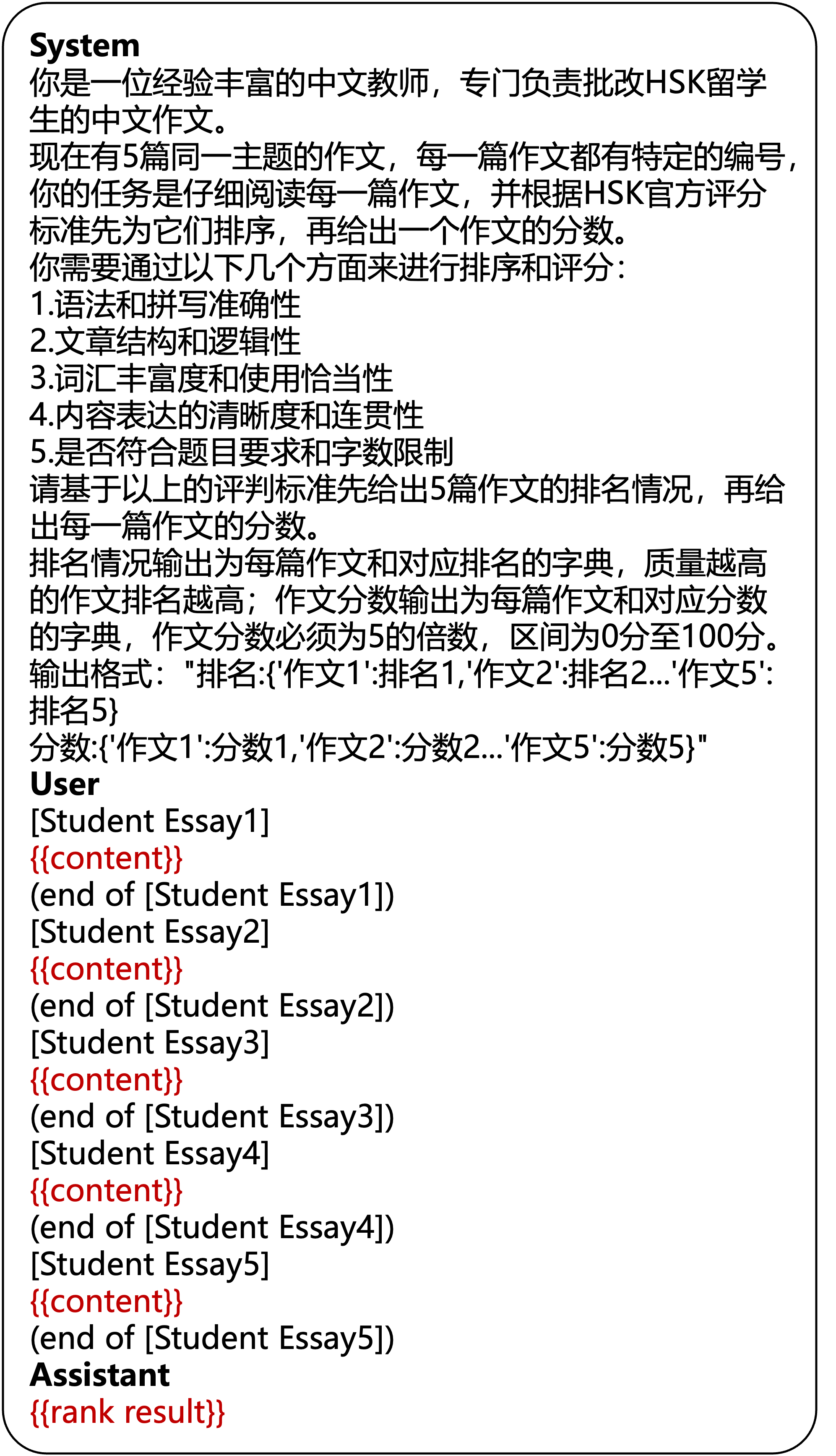}
    \caption{Instruction for the method of scoring in 5 essays. Contents to be filled are highlighted in \textcolor{darkred}{red}.}
\end{figure}

\begin{figure}[t]
    \centering
    \includegraphics[width=\columnwidth]{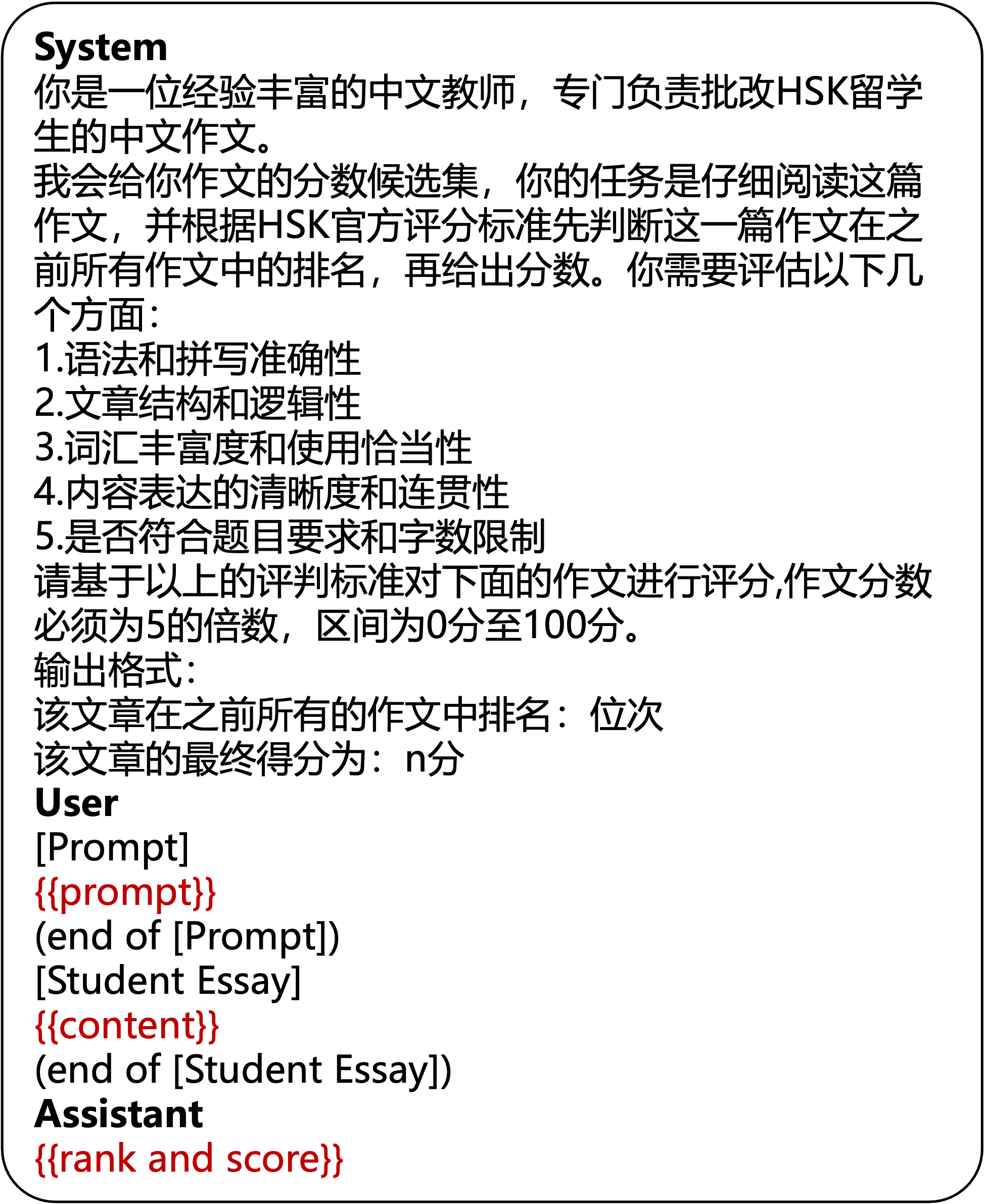}
    \caption{Instruction for the method of simultaneous generation. Contents to be filled are highlighted in \textcolor{darkred}{red}.}
\end{figure}
\clearpage

\newpage
\section{Dataset Description}
\label{sec:hsk_data}
\begin{table}[h]
\centering
\fontsize{11}{14}\selectfont
\begin{tabularx}{\columnwidth}{cXc}
\hline
\textbf{Idx} & \textbf{\#Prompt} & \textbf{Num} \\ \hline
1 & The Impact of Smoking on Personal Health and Public Interest & 1220 \\ 
2 & My Views on Gender-Specific Classes & 340 \\ 
3 & A Job Application Letter & 495 \\ 
4 & Green Food and Hunger & 1402 \\  
5 & Views on "Euthanasia" & 655 \\  
6 & Reflections on "Three Monks Have No Water to Drink" & 894 \\  
7 & The Person Who Influenced Me the Most & 643 \\  
8 & How to Address the "Generation Gap" & 778 \\  
9 & Parents as the First Teachers of Children & 822 \\  
10 & My Views on Pop Music & 704 \\  
11 & A Letter to My Parents & 644 \\ \hline
12 & Athlete Salaries & 36 \\  
13 & The Harm of Silent Environments on the Human Body & 92 \\  
14 & The Joys and Struggles of Learning Chinese & 198 \\  
15 & One of My Holidays & 294 \\  
16 & Views on "Wives Returning Home" & 12 \\  
17 & My Childhood & 183 \\  
18 & The Ideal Way to Make Friends & 228 \\  
19 & My Father & 121 \\  
20 & How to Face Setbacks & 267 \\  
21 & Why I Learn Chinese & 107 \\  
22 & Gum and Environmental Sanitation & 15 \\  
23 & My Views on Divorce & 67 \\  
24 & My Favorite Book & 42 \\  
25 & On Effective Reading & 70 \\ \hline
\end{tabularx}
\caption{The prompts of the HSK dataset are displayed as shown above, with the first 11 prompts utilized for experimentation.}
\label{tab:topics}
\end{table}
\begin{table}[b]
    \centering
    \fontsize{9}{10}\selectfont
    \begin{tabularx}{\columnwidth}{Xcccccc}
        \hline
        \textbf{Dataset} & \textbf{Prompt} & \textbf{\#Essay} & \textbf{Avg Len} & \textbf{Range} & \textbf{Diff} \\ \hline
        \multirow{11}*{HSK} & 1 & 1220 & 355 & 40-100 & 5 \\
                          & 2 & 340 & 434 & 40-100 & 5 \\
                          & 3 & 495 & 353 & 40-100 & 5 \\
                          & 4 & 1402 & 360 & 40-100 & 5 \\
                          & 5 & 655 & 366 & 40-100 & 5 \\
                          & 6 & 894 & 365 & 40-100 & 5 \\
                          & 7 & 643 & 416 & 40-100 & 5 \\
                          & 8 & 778 & 391 & 40-100 & 5 \\
                          & 9 & 822 & 373 & 40-100 & 5 \\
                          & 10 & 704 & 365 & 40-100 & 5 \\
                          & 11 & 644 & 403 & 40-100 & 5 \\ \hline
        \multirow{8}*{ASAP} & 1 & 1783 & 427 & 2-12 & 1 \\
                          & 2 & 1800 & 432 & 1-6 & 1 \\
                          & 3 & 1726 & 124 & 0-3 & 1 \\
                          & 4 & 1772 & 106 & 0-3 & 1 \\
                          & 5 & 1805 & 142 & 0-4 & 1 \\
                          & 6 & 1800 & 173 & 0-4 & 1 \\
                          & 7 & 1569 & 206 & 0-30 & 1 \\
                          & 8 & 723 & 725 & 0-60 & 1 \\ \hline
    \end{tabularx}
\caption{Statistics of two datasets. \textbf{\#Essay} represents the number of essays. \textbf{Avg Len} represents the average number of words. \textbf{Range} represents the score range. \textbf{Diff} represents the common difference.}
\label{tab:dataset_stats}
\end{table}
\clearpage

\begin{figure}[t]
    \centering
    \includegraphics[width=\columnwidth]{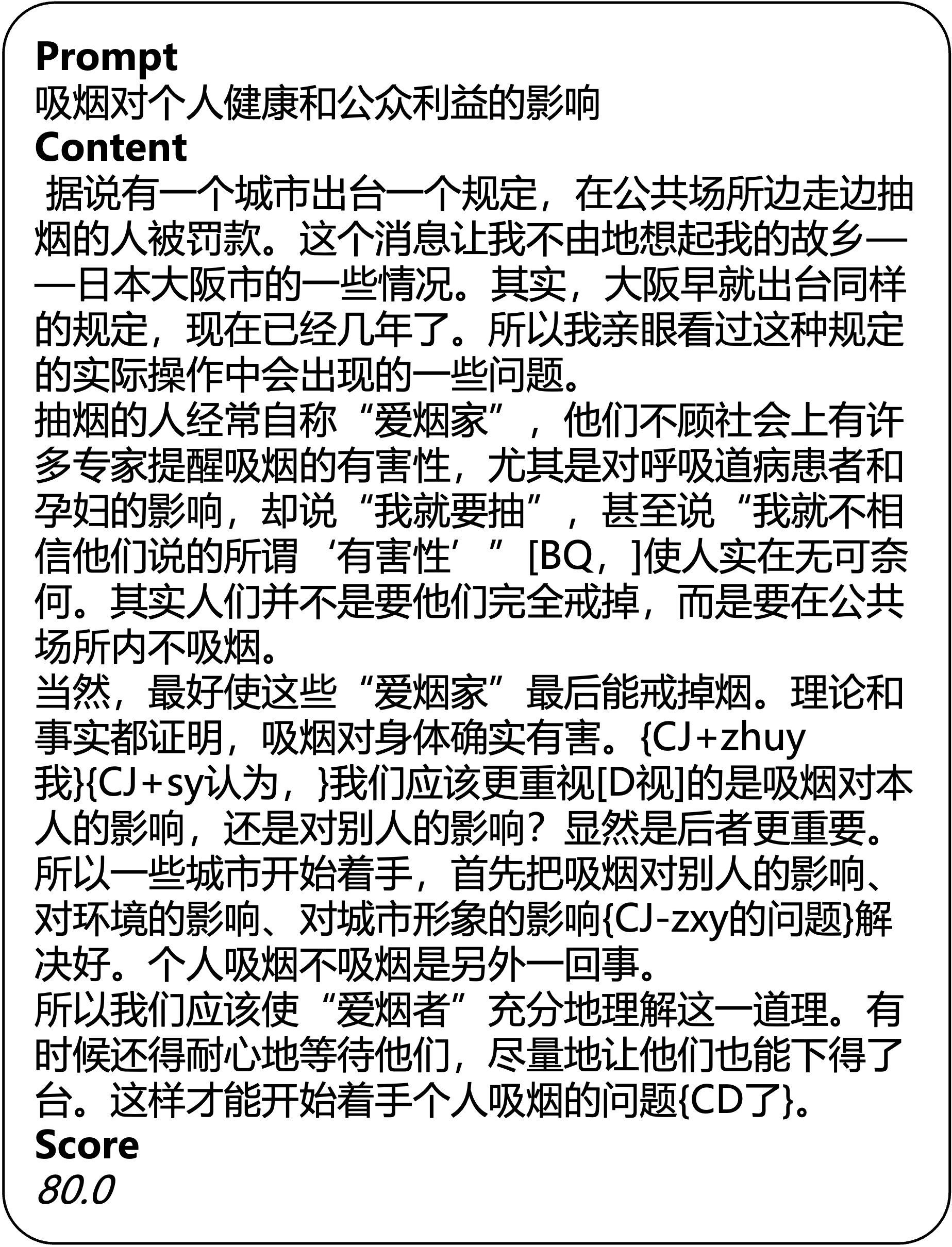}
    \caption{The uncleaned sample essay from the HSK Dataset, which contains flags for syntax errors.}
\end{figure}
\begin{figure}[t]
    \centering
    \includegraphics[width=\columnwidth]{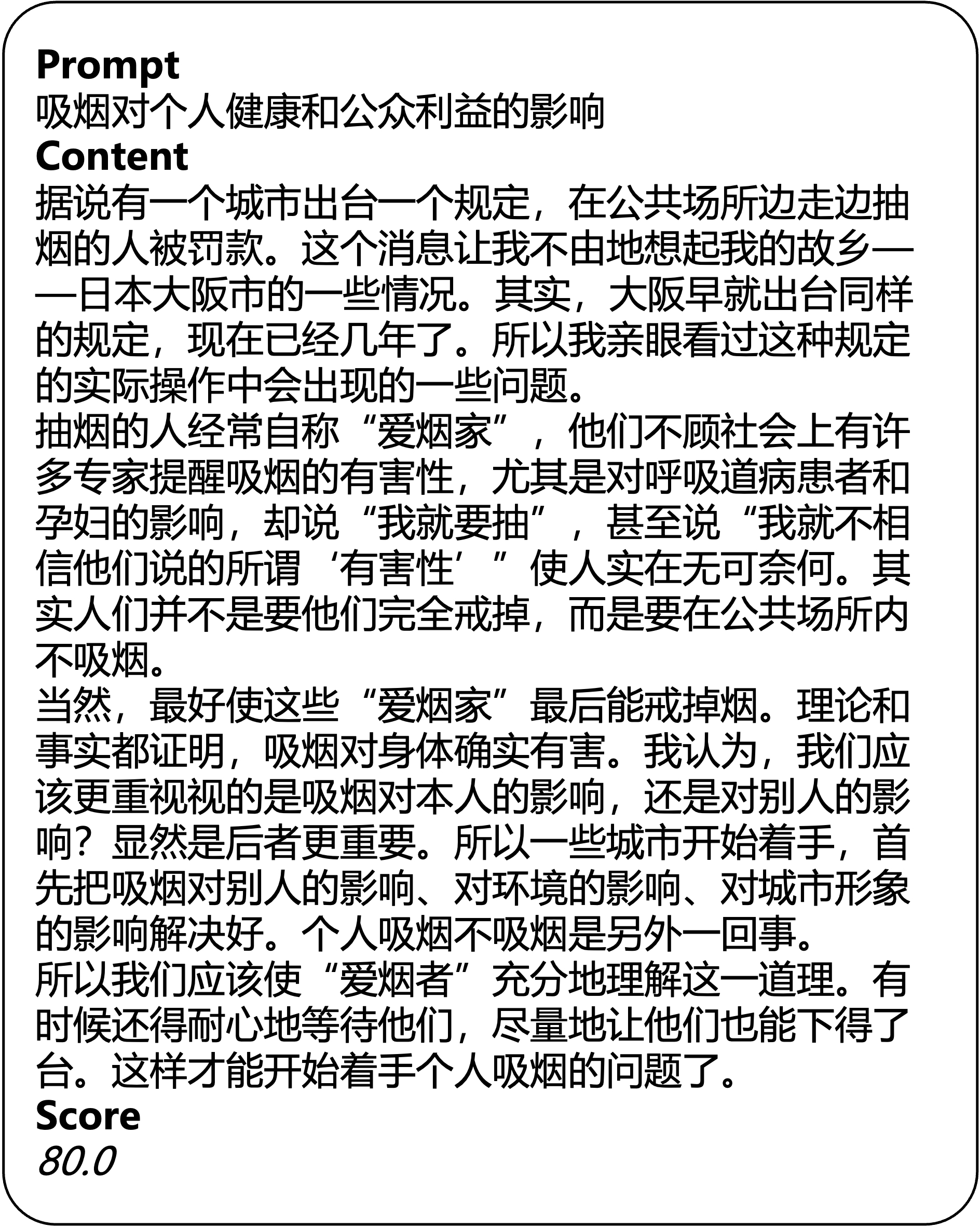}
    \caption{The cleaned sample essay from the HSK Dataset. }
\end{figure}

\end{document}